\documentclass[letterpaper, 10 pt, conference]{ieeeconf}  %
\IEEEoverridecommandlockouts                              %
\usepackage{amsmath} %
\usepackage{amssymb}  %
\usepackage{xcolor}
\usepackage{ifthen} %
\usepackage{multirow}
\usepackage{booktabs}
\usepackage{graphicx}
\usepackage{url}
\usepackage{tabularx}
\usepackage{subcaption}
\usepackage{hyperref} %

\newif\ifshowcomments
\showcommentstrue %

\title{\LARGE \bf
Empirical Analysis of Sim-and-Real Cotraining of Diffusion Policies for Planar Pushing from Pixels
}

\author{Adam Wei, Abhinav Agarwal, Boyuan Chen, Rohan Bosworth, Nicholas Pfaff, Russ Tedrake%
\thanks{
This work was supported by Amazon.com PO\# 2D-06310236, PO\# 2D-15431240, CC OTH 00596792 2021 TR/PO 2D-15694048, The AI Institute, the Toyota Research Institute PO-003816, ONR N00014-22-1-2121, the National Science Foundation Graduate Research Fellowship Program under Grant No. 2141064, and the Natural Sciences and Engineering Research Council of Canada CGS D-587703. Any opinions, findings, and conclusions or recommendations expressed in this material are those of the author(s) and do not necessarily reflect the views of the National Science Foundation. We thank MIT Supercloud \cite{reuther2018interactive} for providing computational resources.
}
\thanks{
All authors are with the Massachusetts Institute of Technology, Cambridge, MA, 02139. Corresponding author: {\tt\small weiadam@mit.edu}}
}

\begin{document}

\maketitle
\thispagestyle{empty}
\pagestyle{empty}

\begin{abstract}
    \emph{Cotraining} with demonstration data generated both in simulation and on real hardware has emerged as a promising recipe for scaling imitation learning in robotics. This work seeks to elucidate basic principles of this \emph{sim-and-real cotraining} to inform simulation design, sim-and-real dataset creation, and policy training. Our experiments confirm that cotraining with simulated data can dramatically improve performance, especially when real data is limited. We show that these performance gains scale with additional simulated data up to a plateau; adding more real-world data increases this performance ceiling. The results also suggest that reducing physical domain gaps may be more impactful than visual fidelity for non-prehensile or contact-rich tasks. Perhaps surprisingly, we find that some visual gap can help cotraining – binary probes reveal that high-performing policies must learn to distinguish simulated domains from real. We conclude by investigating this nuance and mechanisms that facilitate positive transfer between sim-and-real. Focusing narrowly on the canonical task of planar pushing from pixels allows us to be thorough in our study. In total, our experiments span 50+ real-world policies (evaluated on 1000+ trials) and 250 simulated policies (evaluated on 50,000+ trials). Videos and code can be found at \url{https://sim-and-real-cotraining.github.io/}.

\end{abstract}

\section{Introduction}
\label{sec:introduction}

Foundation models trained on large datasets have transformed natural language processing \cite{geminiteam2024geminifamilyhighlycapable}\cite{brown2020languagemodelsfewshotlearners} and computer vision \cite{sun2017revisitingunreasonableeffectivenessdata}. However, this data-driven recipe has been challenging to replicate in robotics since real-world data for imitation learning can be expensive and time-consuming to collect \cite{firoozi2023foundationmodelsroboticsapplications}. Fortunately, alternative data sources, such as simulation and video, contain useful information for robotics. In particular, simulation is promising since it can automate robot-specific data collection. This paper investigates the problem of \textit{cotraining} policies via imitation learning with both simulated and real-world data. Our results confirm that simulation is a powerful tool for scaling imitation learning and improving performance. We also provide insights into the factors and underlying principles that affect \textit{sim-and-real cotraining}.

Many researchers are investing in simulation for data generation in robotics \cite{dalal2023imitatingtaskmotionplanning}\cite{ha2023scalingdistillingdownlanguageguided}\cite{mandlekar2023mimicgendatagenerationscalable}. Concurrently, large consolidated datasets have made real-world data more accessible \cite{embodimentcollaboration2024openxembodimentroboticlearning}\cite{khazatsky2024droidlargescaleinthewildrobot}. Both data sources are valuable, but neither has proven sufficient on its own \cite{firoozi2023foundationmodelsroboticsapplications}. For instance, simulated data is scalable, but requires sim2real transfer; real-world data does not have this issue, but it is more expensive and time-consuming to collect. Thus, understanding how to use both data sources symbiotically could unlock massive improvements in robot imitation learning. 

Given data from simulation and the real-world, \textit{sim-and-real cotraining} aims to train policies that maximize a real-world performance objective. This is a common and important problem in robotics \cite{openai2019solvingrubikscuberobot}\cite{rudin2022learningwalkminutesusing}\cite{ankile2024imitationrefinementresidual}. Our goal is to understand the mechanisms underlying sim-and-real cotraining for imitation learning. We study how various factors affect performance, such as data scale, data mixtures, and distribution shifts. These findings can inform best practices for both sim-and-real cotraining and simulator design for synthetic data generation. 

\begin{figure}[t]
    \centering
    \includegraphics[width=\linewidth]{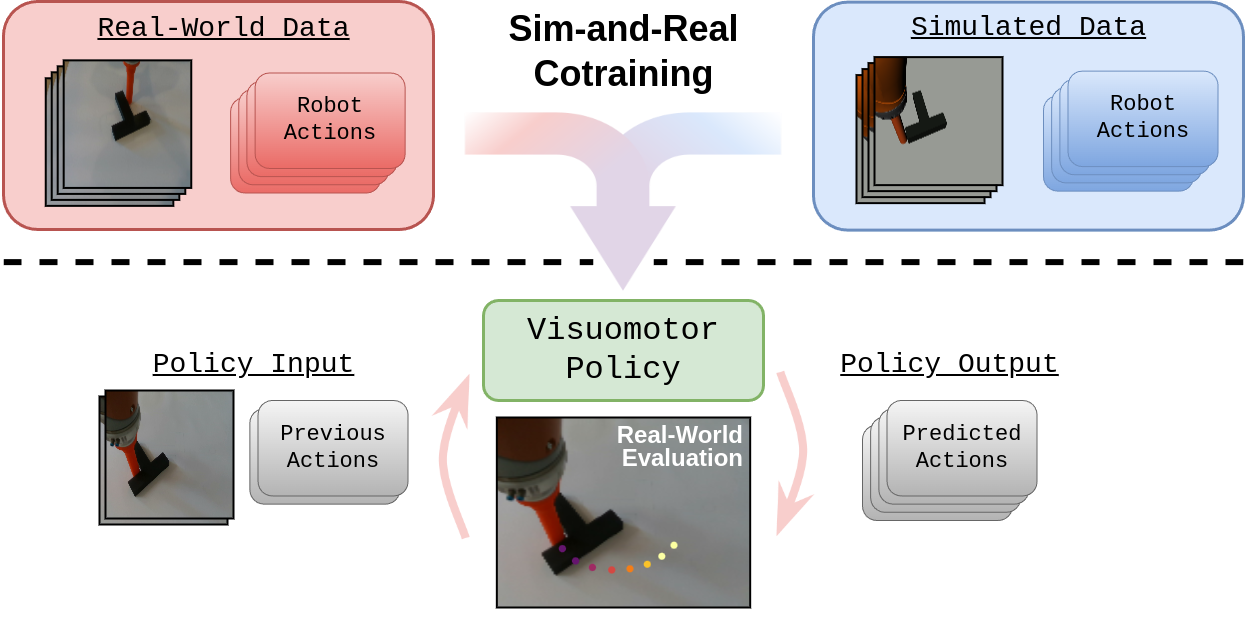}
    \caption{\textit{Sim-and-real cotraining} aims to train visuomotor policies using both simulated and real-world robot data to maximize performance on a real-world objective.\vspace{-1\baselineskip}}
    \label{fig:anchor}
\end{figure}

Specifically, we investigate sim-and-real cotraining for Diffusion Policy \cite{chi2024diffusionpolicy} and evaluate performance on the canonical task of planar-pushing from pixels. We provide an extensive suite of experiments and ablations that spans over 50 real-world policies (evaluated on 1000+ trials) and 250 simulated policies (evaluated on 50,000+ trials). More concretely, we show that:
\begin{enumerate}
    \item \textbf{Cotraining with sim data improves policy performance by up to 2-7x}, but the performance gains from scaling sim eventually plateau without more real data.
    \item All sim2real gaps impact the value of synthetic data. In particular, \textbf{improving the physical accuracy of simulators} could massively increase the downstream performance of cotrained policies.
    \item \textbf{High-performing policies learn to distinguish sim from real} since the physics of each environment require different actions. Surprisingly, cotraining with perfectly rendered sim data reduces performance since the policies can no longer visually discern the two domains. In contrast, providing a one-hot encoding of the environment improves policy performance.
    \item \textbf{Cotraining from sim provides positive transfer to real.} Our results show that simulated data can fill gaps in the real-world data, and scaling sim decreases the real-world test loss according to a power law.
    \item We also provide \textbf{ablations for two alternative cotraining formulations}: one that uses classifier-free guidance \cite{ho2022classifierfreediffusionguidance} and another that modifies the loss function to encourage better representation learning.
\end{enumerate}

Focusing on the single canonical task of planar pushing from pixels allowed us to be exhaustive in our investigation, but it also limits our study. Thus, this paper's main contributions are the trends and analysis rather than the absolute values. We hope our findings can inform larger-scale cotraining, where thorough sweeps and analysis could be prohibitively expensive.

\section{Related Work}
\label{sec:related_work}
\subsubsection{Data Generation in Simulation}
\label{related_works:data_generation}
Prior works have shown that simulation can scale up data generation \cite{dalal2023imitatingtaskmotionplanning}\cite{ha2023scalingdistillingdownlanguageguided}\cite{zhu2024learncontactrichmanipulationpolicies} and augmentation \cite{mandlekar2023mimicgendatagenerationscalable}\cite{yang2025physics} for robot imitation learning. Additionally, infrastructure for synthetic data generation is improving, with advancements in environment generation \cite{torne2024reconcilingrealitysimulationrealtosimtoreal}, real2sim \cite{pfaff2025scalablereal2simphysicsawareasset}, and motion planning \cite{marcucci2022motionplanningobstaclesconvex}. This paper investigates how the growing body of work in simulated data generation can be used in conjunction with real-world datasets \cite{embodimentcollaboration2024openxembodimentroboticlearning}\cite{khazatsky2024droidlargescaleinthewildrobot} for robot imitation learning.

\subsubsection{Sim2real Transfer vs Sim-and-Real Cotraining}
\label{related_works:sim_real_cotraining}
Sim-and-real cotraining adopts a slightly different philosophy to policy learning than the traditional \textit{sim2real} pipeline \cite{rudin2022learningwalkminutesusing}\cite{chen2021generalinhandobjectreorientation}. Instead of learning policies in sim and applying sim2real techniques \cite{openai2019solvingrubikscuberobot}\cite{tobin2017domainrandomizationtransferringdeep}\cite{fey2025bridgingsimtorealgapathletic}, robots learn from sim and real-world data \textit{simultaneously}. Prior works have used sim-and-real cotraining on assembly \cite{ankile2024imitationrefinementresidual} and kitchen tasks \cite{nasiriany2024robocasalargescalesimulationeveryday}.

The general problem of learning from multiple domains is ubiquitous in machine learning \cite{geminiteam2024geminifamilyhighlycapable}\cite{li2021universalrepresentationlearningmultiple}\cite{theorydifferentdomains}\cite{xie2023doremioptimizingdatamixtures} In robotics, most works have focused on cotraining from \textit{real} domains \cite{embodimentcollaboration2024openxembodimentroboticlearning}\cite{kim2024openvlaopensourcevisionlanguageactionmodel}\cite{hejna2024remixoptimizingdatamixtures}; the specific problem of cotraining from simulated domains can be seen as a special case of cross-embodiment training worthy of focused study. We believe simulation will play a key role in robot imitation learning given the progress in synthetic data generation and the ability to perform large-scale reproducible testing. This work aims to provide initial insights towards this future.

\section{Preliminaries}
\label{sec:preliminaries}

\subsection{Cotraining Problem Formulation and Notation}
\label{preliminaries:problem_formulation}
We study a specific instantiation of the cotraining problem, where a behavior cloning policy is jointly trained on simulated and real-world robot data. Let $\tau = \{(\boldsymbol{o}_i, \boldsymbol{a}_i)\}_{i=1}^{L}$ be a trajectory of observation-action pairs. Let $\mathcal D_R = \{\tau_i\}_{i=1}^{N_R}$ and $\mathcal D_S = \{\tau_i\}_{i=1}^{N_S}$ be datasets of real and sim trajectories respectively. Let $|\mathcal D|$ be the number of trajectories in $\mathcal D$. Given $\mathcal D_R$ and $\mathcal D_S$, cotraining aims to learn a policy that maximizes performance on a real-world performance objective. Here, we cotrain Diffusion Policies \cite{chi2024diffusionpolicy} and measure performance as the binary success rate on planar-pushing from pixels.

\subsection{Diffusion Policy}
\label{preliminaries:diffusion_policy}
Diffusion Policies sample future robot actions given a history of past observations to accomplish a desired task \cite{chi2024diffusionpolicy}. More concretely, they learn a denoiser for the conditional action distribution $p(\mathbf{A}|\mathbf{O})$ by minimizing the loss in \eqref{eq:denoiser_loss}. $\mathbf{A}$ and $\mathbf{O}$ are \emph{horizons} of actions and observations, $\rho_t$ and $\sigma_t$ are parameters of the noise schedule, and $\theta$ are the parameters of the denoiser. We sample from the learned policy, $\pi_\theta(\mathbf{A}|\mathbf{O})$, by iteratively applying $\epsilon_\theta$ in a denoising process \cite{song2022denoisingdiffusionimplicitmodels}\cite{ho2020denoisingdiffusionprobabilisticmodels}.
\begin{equation}
    \mathcal L_\mathcal D(\theta) = \mathbb E_{t,(\mathbf{O},\mathbf{A})\sim \mathcal D,\epsilon\sim \mathcal N(0,I)}[\lVert \epsilon - \epsilon_\theta(\rho_t\mathbf{A} + \sigma_t\epsilon,\ \mathbf{O},\ t) \rVert^2_2]
\label{eq:denoiser_loss}
\end{equation}
Our experiments train Diffusion Policies since they are a state-of-the-art algorithm for imitation learning \cite{chi2024diffusionpolicy}. We use ResNet18 \cite{he2015deepresiduallearningimage} as the vision backbone and train end-to-end. Training details are available in Appendix \ref{appendix:training}.

\subsection{Planar-Pushing From Pixels}
\label{preliminaries:planar_pushing}

\begin{figure}[t]
    \centering
    \includegraphics[width=0.9\linewidth, trim={0 1.75cm 0 1.75
    cm}, clip]{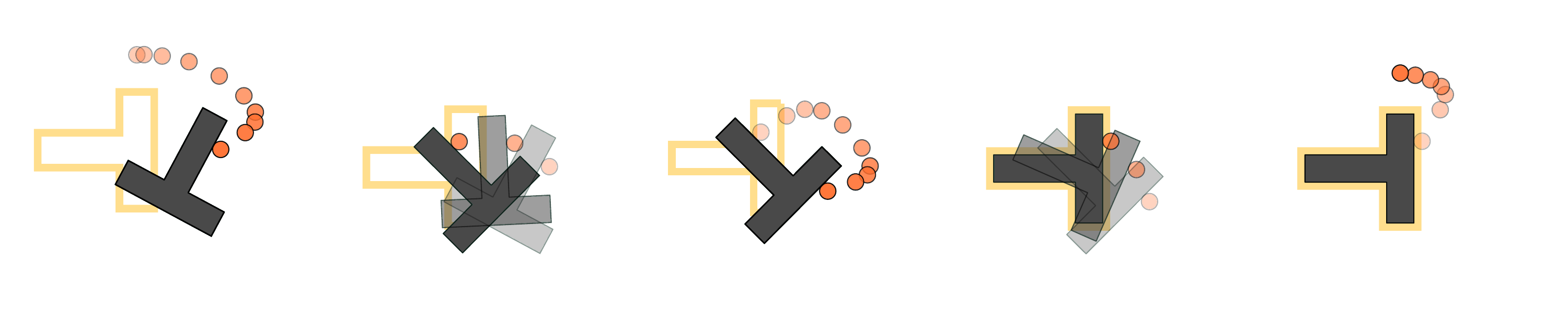}
    \caption{An example of planar-pushing \cite{graesdal2024tightconvexrelaxationscontactrich}. The circle is the pusher, the black T is the slider, and the yellow T is the goal.\vspace{-\baselineskip}}
    \label{fig:planar_pushing}
\end{figure}

Planar-pushing is a manipulation problem that requires a robot to push an object (slider) on a flat surface with a cylindrical end effector (pusher) to a target pose (see Figure \ref{fig:planar_pushing}). \textit{``From pixels''} indicates that the observation space for the task contains images as opposed to the full system state.

We emphasize that the goal of this work is not to ``solve'' planar pushing. Instead, planar pushing serves as a test-bed for our investigations into cotraining. We chose this task since it is a canonical task that captures core challenges in robotics \cite{chi2024diffusionpolicy}\cite{lee2024behaviorgenerationlatentactions}\cite{lynch2022interactivelanguagetalkingrobots}, such as high-level reasoning, visuomotor control, and contact, while still admitting methods for data generation \cite{graesdal2024tightconvexrelaxationscontactrich}\cite{aydinoglu2024consensuscomplementaritycontrolmulticontact}\cite{kang2025globalcontactrichplanningsparsityrich}.

\subsection{Data Collection}
\label{preliminaries:data_collection}
A human teleoperator collects real data and an optimization-based planner \cite{graesdal2024tightconvexrelaxationscontactrich} generates simulated trajectories. Each sim trajectory is replayed in {\tt Drake} \cite{tedrake2019drake} to render the observations. The simulated planner is ``near-optimal'' \cite{graesdal2024tightconvexrelaxationscontactrich}, whereas the human teleoperator is not. This introduces an action gap between the two datasets that will become relevant for our analysis in Section \ref{sec:analysis}. Lastly, we highlight the robustness of our pipeline. In this paper, we generated over 25,000 \textit{unique} trajectories without any cost-tuning.

The simulation environment mimics the real-world setup. Both datasets are collected on a {\tt KUKA LBR iiwa 7}. The action space is the target $x$-$y$ position of the pusher (the robot's end effector); the pusher’s $z$-value and orientation are fixed. The observation space includes the pusher's pose and two RGB images from an overhead camera and a wrist camera. Figure \ref{fig:anchor} visualizes the overhead camera view.

\section{Real World Experiments}
\label{sec:real_world}
\begin{figure*}[t]
    \centering
    \includegraphics[width=0.32\linewidth]{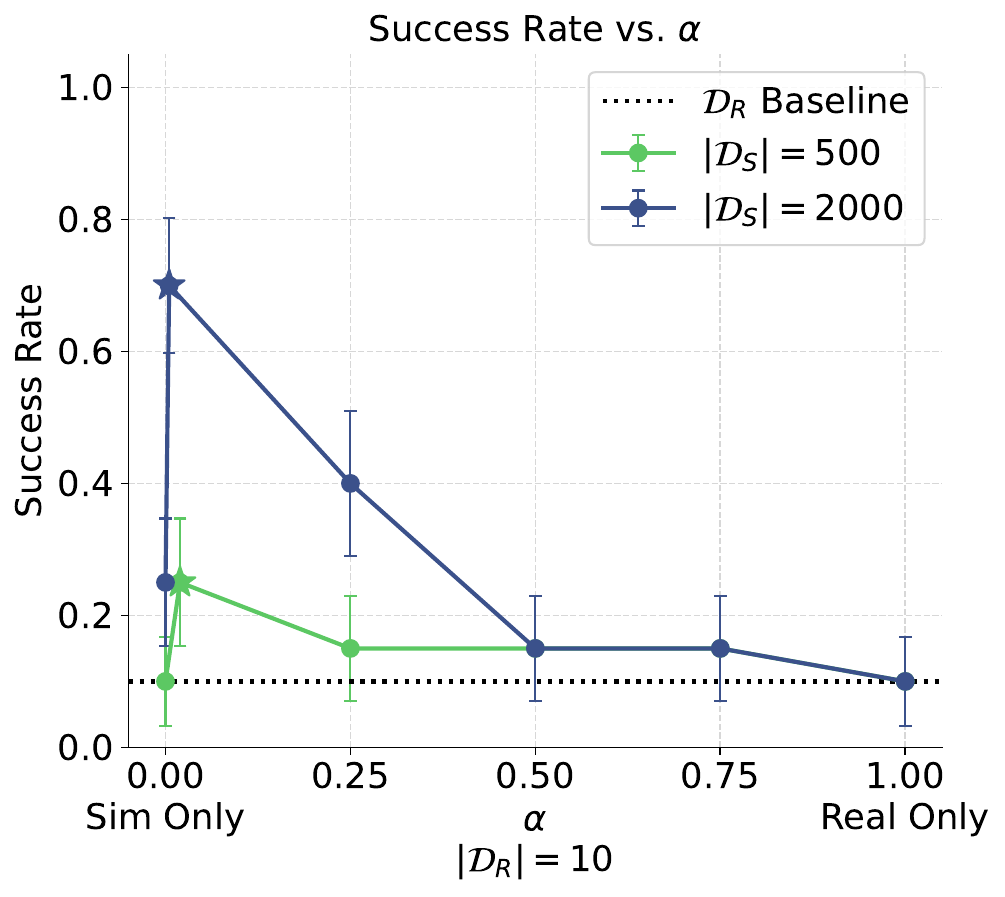}
    \includegraphics[width=0.32\linewidth]{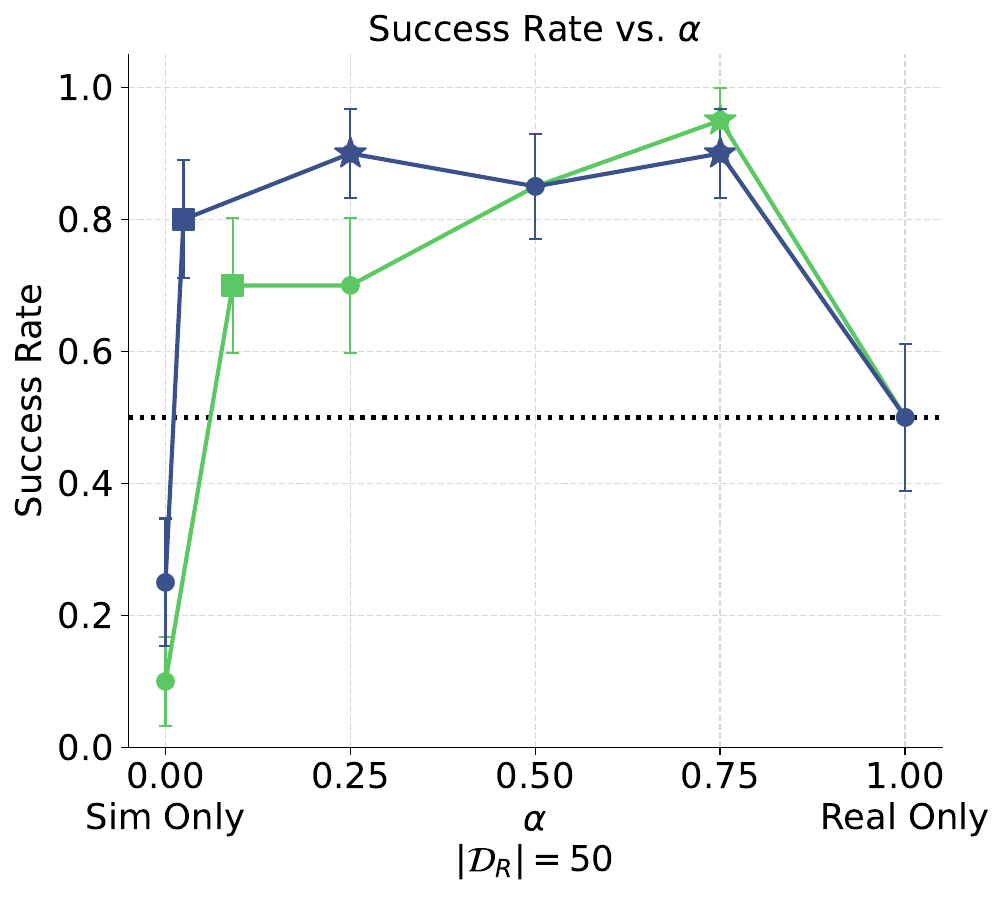}
    \includegraphics[width=0.32\linewidth]{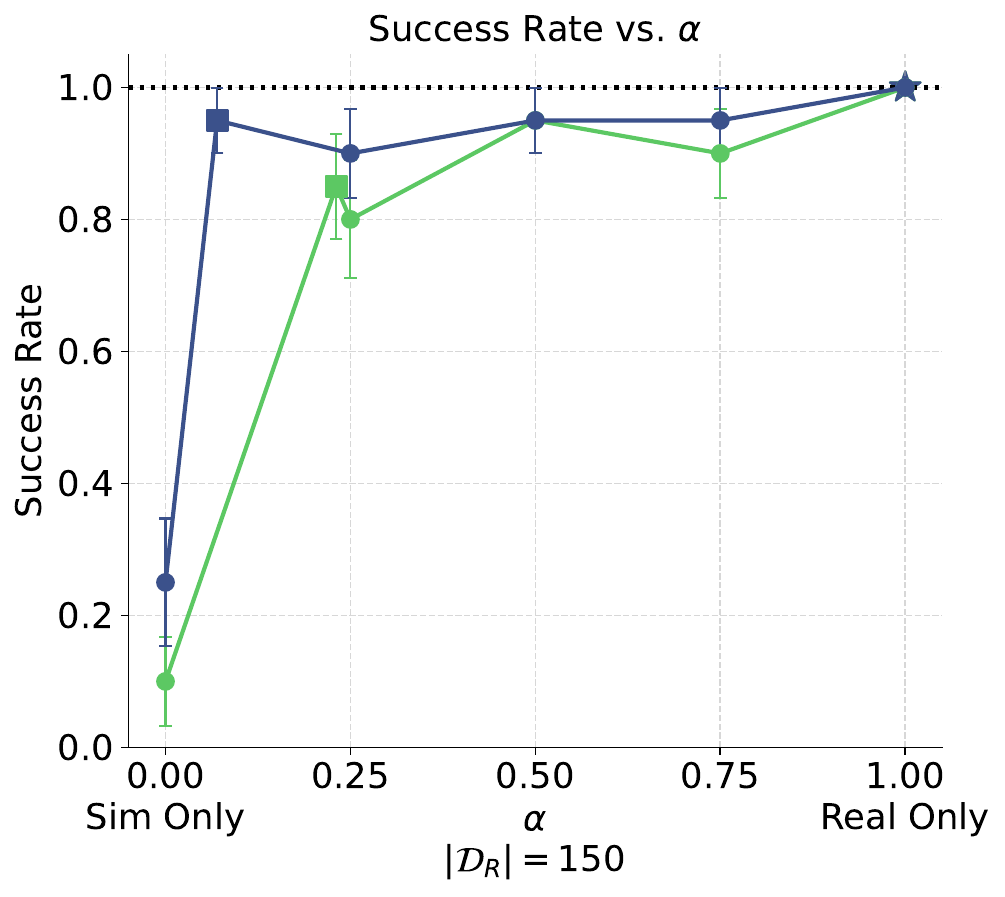}
    \caption{Real-world performance of cotrained policies at different data scales and mixing ratios. \scalebox{0.7}{$\bigstar$} depicts the optimal $\alpha$ and \scalebox{0.6}{$\blacksquare$} depicts the natural mixing ratio. When real data is limited, cotraining with sim data can improve performance by 2-7x.}  \vspace{-\baselineskip}
    \label{fig:real_world_cotraining_results}
\end{figure*}

One way to cotrain from multiple domains is to mix or reweight the datasets \cite{kim2024openvlaopensourcevisionlanguageactionmodel}\cite{theorydifferentdomains}. We investigate the following questions about this simple, but common, algorithm:
\begin{enumerate}
    \item Does cotraining with sim data improve real-world policy performance?
    \item How do the size and the mixing ratio between both datasets affect real-world performance?
    \item How does cotraining compare with finetuning?
\end{enumerate}

\subsection{Experimental Setup}
\label{real_world:experiment}
We refer to the method above as \textit{vanilla cotraining}. In vanilla cotraining, policies are trained on $\mathcal D^\alpha$, a mixture of the real data, $\mathcal D_R$, and sim data, $\mathcal D_S$. We sample from $\mathcal D^\alpha$ by sampling from $\mathcal D_R$ with probability $\alpha$ and $\mathcal D_S$ otherwise. By the tower property of expectations, training on $\mathcal D^\alpha$ is equivalent to minimizing \eqref{eq:cotraining_loss}. Thus, $\alpha$ acts as both a mixing ratio and a ``reweighting'' parameter. 
\begin{equation}
    \mathcal{L}_{\mathcal{D}^{\alpha}} = \alpha \mathcal{L}_{\mathcal D_R} + (1 - \alpha) \mathcal{L}_{\mathcal D_S}
\label{eq:cotraining_loss}
\end{equation}

Ideally, $\mathcal L_{\mathcal D_R}$ ($\alpha=1$) is our training objective; however, when $|\mathcal D_R|$ is small, this leads to overfitting and poor performance. We remedy this by adding sim data and using $\alpha$ to prevent $\mathcal D_S$ from dominating the loss when $|\mathcal D_S| \gg |\mathcal D_R|$. Historically, considerable resources are used to sweep dataset reweightings (or equivalently $\alpha$ in our setting) since they have an outstanding effect on performance \cite{geminiteam2024geminifamilyhighlycapable}\cite{kim2024openvlaopensourcevisionlanguageactionmodel}.

We cotrain policies for all combinations of $|\mathcal D_R|\ $$\in\ $$\{10, 50, 150\}$, $|\mathcal D_S|\ $$\in\ $$\{500, 2000\}$, and $\alpha\ $$\in\ $$\{0, \tfrac{|D_R|}{|D_R| + |D_S|}, 0.25, 0.5, 0.75, 1\}$. $\alpha=0$ and $\alpha=1$ are equivalent to training solely on $\mathcal D_S$ or $\mathcal D_R$ respectively (i.e no cotraining); $\alpha=\tfrac{|\mathcal D_R|}{|\mathcal D_R| + |\mathcal D_S|}$ is the natural mixing ratio since it is equivalent to concatenating $\mathcal D_R$ and $\mathcal D_S$ without reweighting. We evaluate each policy's real-world success rate and error bars according to Appendix \ref{appendix:real}. The error bars capture the performance distribution of the best checkpoint, but \emph{do not} capture variability due to stochasticity in training.

\subsection{Does Cotraining Improve Performance?}
\label{real_world:results}
Figure \ref{fig:real_world_cotraining_results} presents the results. As $|\mathcal D_R|$ increased, the \textit{real-only baslines} (policies trained on only $\mathcal D_R$) achieved success rates of 2/20, 10/20, and 20/20. Thus, we refer to the three values of $|\mathcal D_R|$ as the low, medium, and high data regimes.

In the low and medium data regimes, cotraining \textit{improves performance over the real-only baselines for all $\alpha$'s}. In fact, the best cotrained policy for $|\mathcal D_R|=10$ improved performance from 2/20 to 14/20, and the best cotrained policy for $|\mathcal D_R|=50$ improved performance from 10/20 to 19/20. To achieve similar improvements with only real data, we would have needed to increase $|\mathcal D_R|$ by several times (see Figure \ref{fig:cotraining_plateaus}).

In the high data regime, the real-only baseline achieved a success rate of 20/20; however a perfect score does not imply a perfect policy since our experiments have variance. With this in mind, the performance decrease from cotraining in the high-data regime is minor and within the error margins.

\begin{figure}[t]
    \centering
    \includegraphics[width=0.49\linewidth]{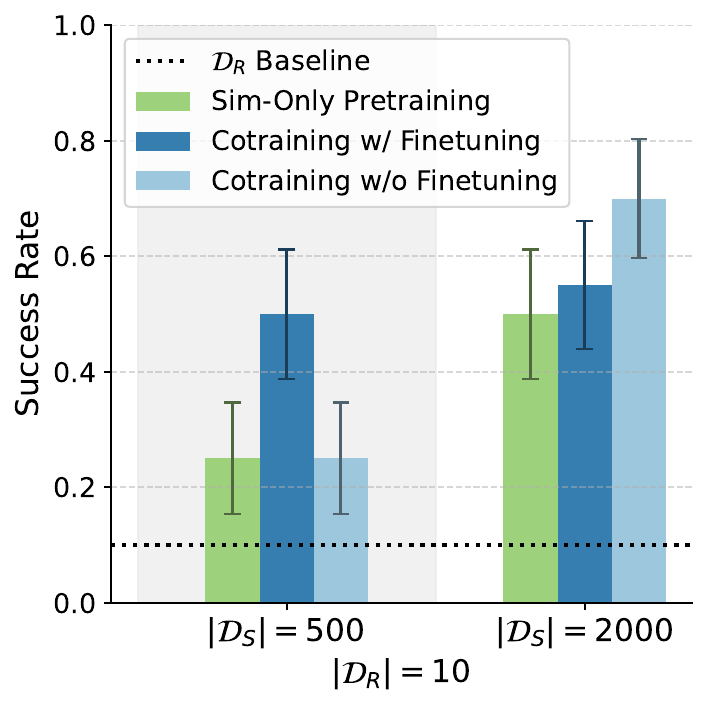}
    \includegraphics[width=0.49\linewidth]{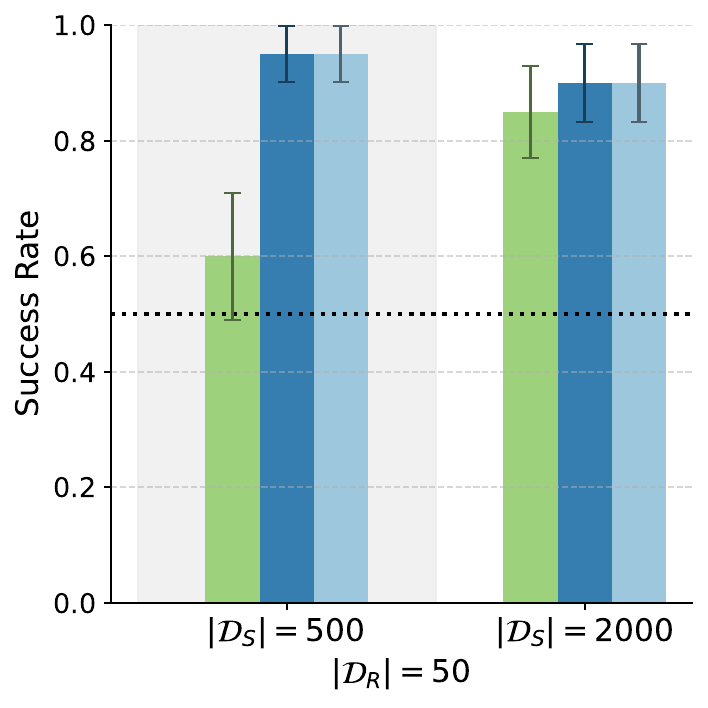}
    \caption{Pretraining with a cotraining mixture significantly outperforms pretraining with sim-only.\vspace{-0.5\baselineskip}}
\label{fig:finetuning}
\end{figure}

\subsection{The Effect of $\alpha$, $|\mathcal D_R|$, and $|\mathcal D_S|$ on Cotraining}
\label{real:effects}
\subsubsection{Effect of $\alpha$}
Performance is sensitive to $\alpha$, especially for $|\mathcal D_R|=10$. The optimal $\alpha$'s appear to increase with $|\mathcal D_R|$, but this trend becomes less pronounced when more sim data is added. Intuitively, overfitting to $\mathcal L_{\mathcal D_R}$ is less detrimental when $|\mathcal D_R|$ is large, so it is desirable to bias the mixing ratio towards real. Performance decreases nearly discontinuously as $\alpha\searrow 0$. This suggests that even a small mixture of real data can drastically improve performance. 

\subsubsection{Effect of $|\mathcal D_R|$}
Cotraining appears most effective in the low to medium data regime. For $|\mathcal D_R|=10$, the best cotrained policy improves performance by 7x, but does not attain near-perfect performance. The real only baseline for $|\mathcal D_R|=50$ performs poorly, but the best cotrained policy performs exceptionally well (19/20). We reached the high-data regime for planar-pushing from pixels with just 150 real demos; however, other tasks often remain in the medium data regime even with thousands of demos \cite{zhao2024alohaunleashedsimplerecipe}. Thus, in most settings, we expect the value of cotraining to be high.

\subsubsection{Effect of $|\mathcal D_S|$}
Scaling $|\mathcal {D}_S|$ improved or maintained performance across the board. Scaling $|\mathcal {D}_S|$ also reduced the policy's sensitivity to $\alpha$, which is highly desirable.

These findings show that simulated data generation and cotraining are promising ways to scale up imitation learning. We verify these results at a larger-scale in Section \ref{sim:cotraining_asymptotes}.

\subsection{Finetuning Comparison}
\label{real_world:finetuning}
We compare cotraining with two methods for finetuning. \textit{1) Sim-only pretraining:} pretrain with $\mathcal D_S$, then finetune with $\mathcal D_R$. \textit{2) Cotraining with finetuning:} cotrain with $\mathcal D_S$, $\mathcal D_R$, and optimal $\alpha$, then finetune with $\mathcal D_R$. Figure \ref{fig:finetuning} shows that methods that employ data mixing outperform methods that train on sim-and-real separately. In the single-task setting, finetuning the cotrained models does not consistently improve performance. We hypothesize that the real-world is already in distribution for the cotrained policies; thus, finetuning could cause overfitting on a case-by-case basis. Results may differ for multi-task sim-and-real cotraining.

\section{Simulation Experiments}
\label{sec:sim}

\begin{figure}[t]
    \centering
    \includegraphics[width=\linewidth]{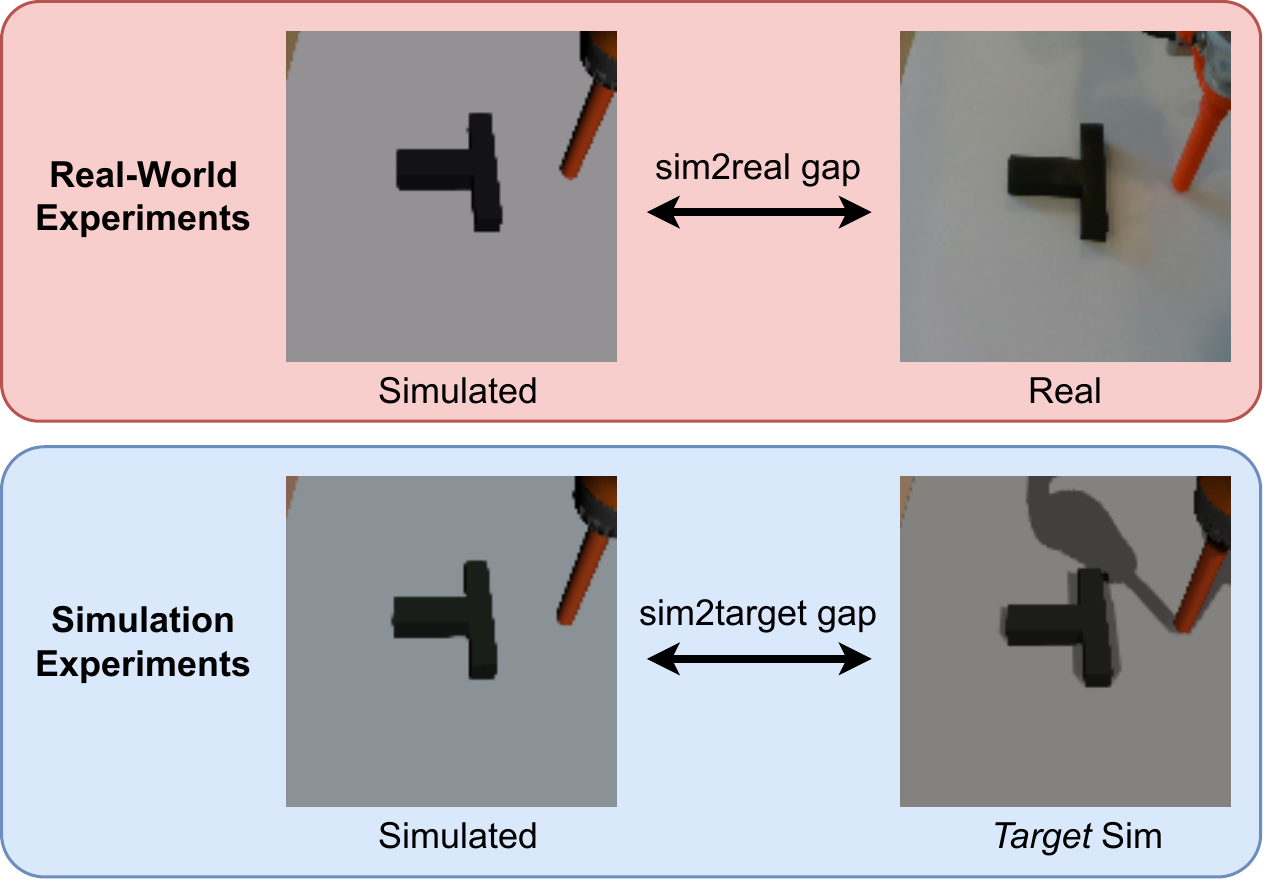}
    \caption{A visualization and comparison of the \textit{sim2real} gap in Section \ref{sec:real_world} and the \textit{sim2target} gap in Section \ref{sec:sim}.\vspace{-\baselineskip}}
    \label{fig:sim_sim}
\end{figure}

In this section, we conduct experiments in simulation on a \textit{sim-and-sim} setup that mimics the \textit{sim-and-real} setup. Simulation enables experiments that would be challenging to conduct in the real-world by providing two main advantages:
\begin{itemize}
    \item\textit{Advantage 1:} automated, high-confidence evaluations.
    \item \textit{Advantage 2:} explicit control over the \textit{sim2real} gap between the two cotraining environments.
\end{itemize}
Section \ref{sim:cotraining_asymptotes} uses \textit{Advantage 1} to scale up the real-world experiments. Section \ref{sim:distribution_shifts} uses \textit{Advantage 2} to study the effect of distribution shifts on cotraining.

\subsection{Sim-and-Sim Setup}
\label{sim:sim-sim}
Cotraining from sim-and-sim requires 2 distinct simulation environments. We reuse the same sim environment from our real-world experiments and continue referring to it is as \textit{sim}. We introduce a second \textit{target} sim environment as a surrogate for the \textit{real-world} environment. Given data from both the \textit{sim} environment and the \textit{target sim} environment, we aim to learn policies that maximize performance in the \textit{target} simulation.

To ensure our simulation experiments are informative for the real world, we designed the \textit{target} sim environment to mimic the sim2real gap (see Table \ref{tab:sim2sim_gap} and Figure \ref{fig:sim_sim}). We collect data in both environments according to Section \ref{preliminaries:data_collection}. Although the \textit{target} dataset serves the same purpose as $\mathcal D_R$ in our real-world experiments, we denote it by $\mathcal D_T$ to make the distinction clear. We evaluate policies in simulation as per Appendix \ref{appendix:sim}. Notably, we test each policy on 200 random trials instead of the 20 used in Section \ref{sec:real_world}.

\begin{table}[t]
    \centering
    \renewcommand{\arraystretch}{1} %
    \begin{tabular}{|>{\centering\arraybackslash}p{0.69cm}|
                    >{\centering\arraybackslash}p{2.2cm}|
                    >{\centering\arraybackslash}p{2cm}|
                    >{\centering\arraybackslash}p{2cm}|}
        \hline
        \textbf{Gap} & \textbf{Simulation} & \textbf{Real World} & \textbf{\textit{Target} Sim} \\ \hline
        Visual & white light, camera \& color offset  
               & natural light, shadows 
               & warm light, shadows \\ \hline
        Physics & Quasistatic 
                & Real physics  
                & {\tt Drake} physics \\ \hline
        Action & Motion planning
               & Human teleop 
               & Human teleop \\ \hline
    \end{tabular}
    \caption{The \textit{sim2target} gap was designed to mimic the \textit{sim2real} gap along 3 different axes.\vspace{-\baselineskip}}
    \label{tab:sim2sim_gap}
\end{table}

\subsection{Single-Task Cotraining Asymptotes}
\label{sim:cotraining_asymptotes}
We scale up the experiments from Section \ref{sec:real_world} in our sim-and-sim setup to answer the following research questions:
\begin{enumerate}
    \item Can we verify our real-world results with higher-confidence evaluations?
    \item How does performance scale with $|\mathcal D_S|$?
\end{enumerate}

We repeat Section \ref{sec:real_world} with $|\mathcal D_T|\in\{10, 50, 150\}$, $|\mathcal D_S|\in\{100, 250, 500, 2000, 4000\}$, and the same $\alpha$'s. We add $\alpha=0.1$ when $\tfrac{|\mathcal D_T|}{|\mathcal D_T|+|\mathcal D_S|}$ is small or close to 0.25. Our finetuning comparison sweeps $|\mathcal D_T|\in\{10, 50\}$ and $|\mathcal D_S|\in\{100, 250, 500, 2000\}$.

\begin{figure*}[t]
    \centering
    \includegraphics[width=0.32\linewidth]{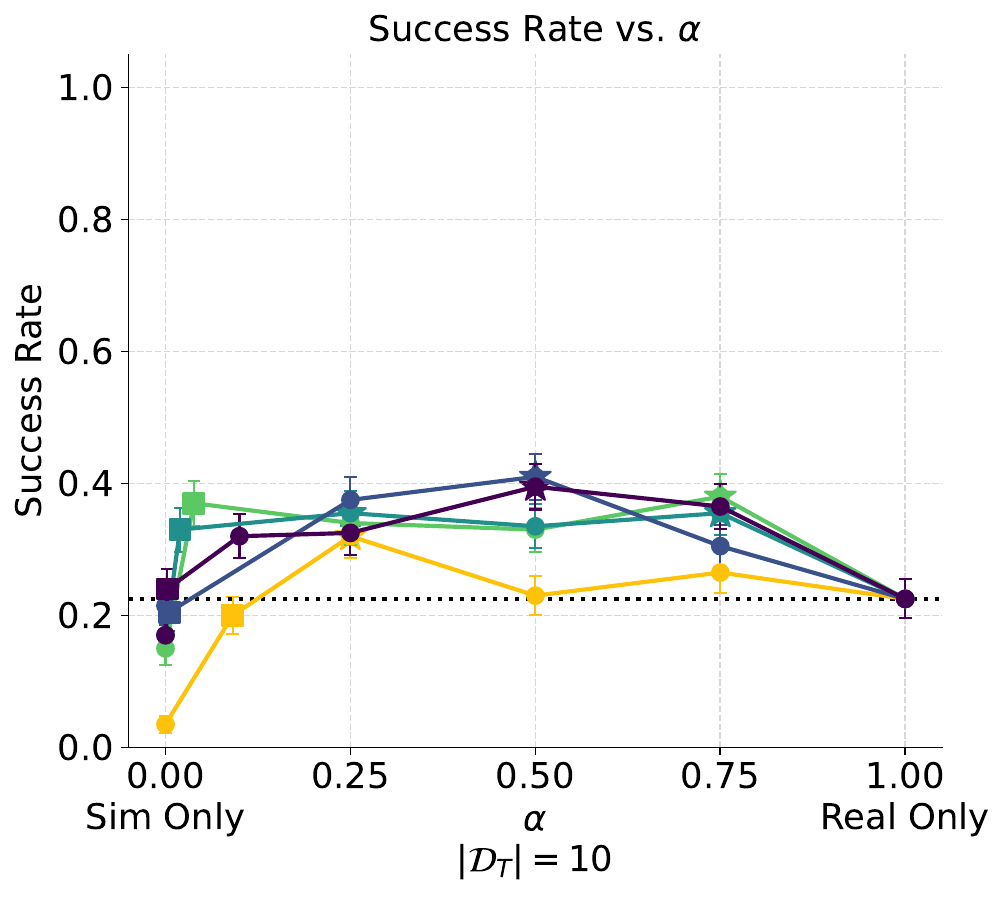}
    \includegraphics[width=0.32\linewidth]{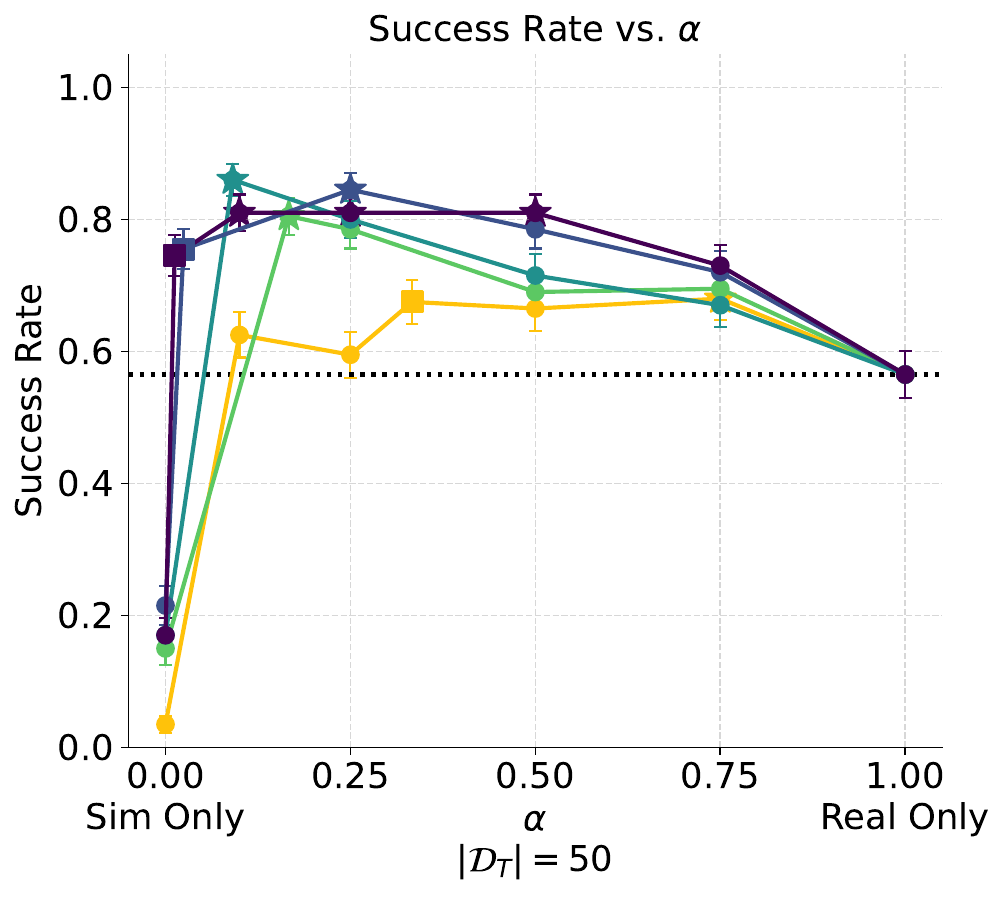}
    \includegraphics[width=0.32\linewidth]{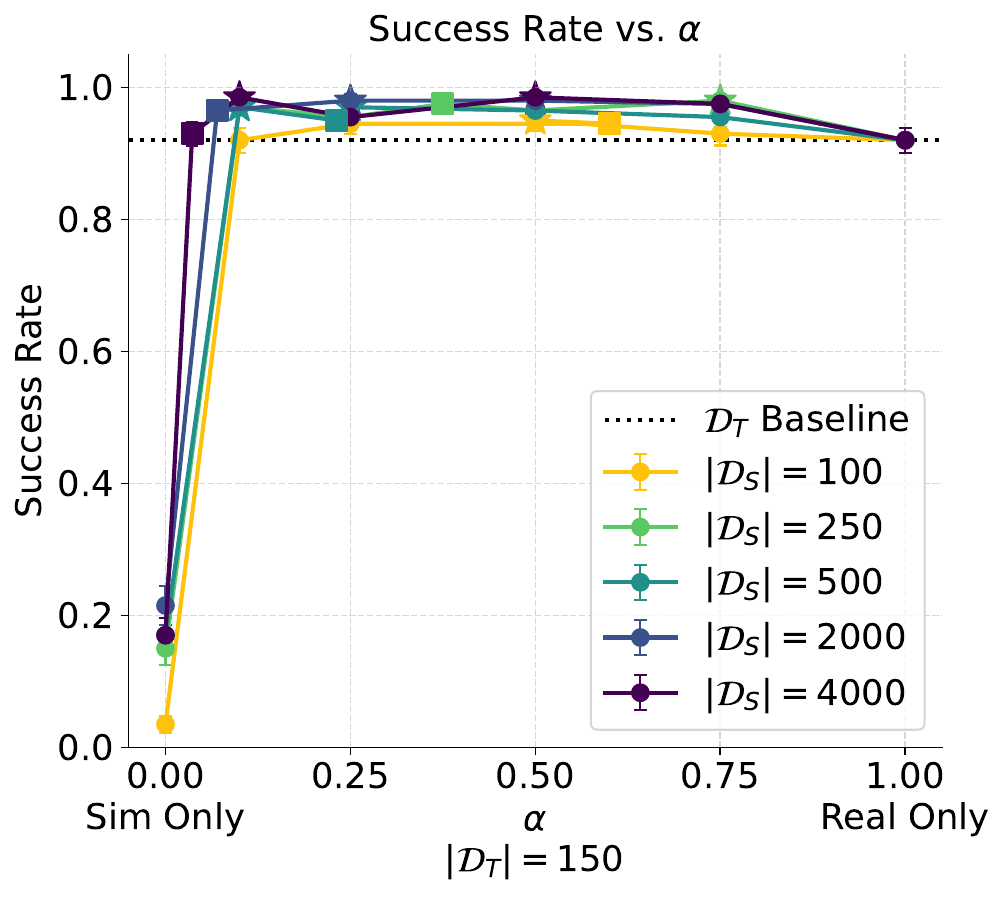}
    \caption{Performance of cotrained policies in simulation at different data scales and mixing ratios. \scalebox{0.7}{$\bigstar$} depicts the optimal $\alpha$ and \scalebox{0.6}{$\blacksquare$} depicts natural mixing ratio. The results of the simulated experiments agree with the real-world results. Scaling up $|\mathcal D_S|$ improves performance until a performance plateau. This plateau can be improved by adding more \textit{target} data.\vspace{-\baselineskip}}
    \label{fig:scaled_cotraining}
\end{figure*}

\begin{figure}[ht]
    \centering
    \includegraphics[width=\linewidth]{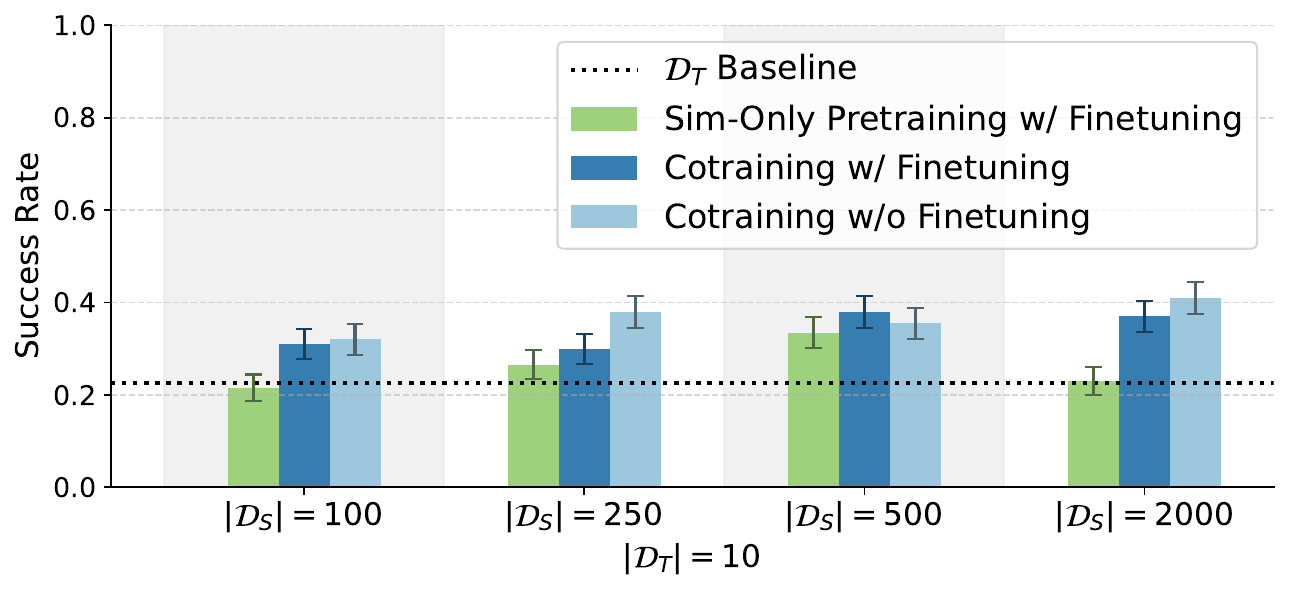}\\
    \includegraphics[width=\linewidth]{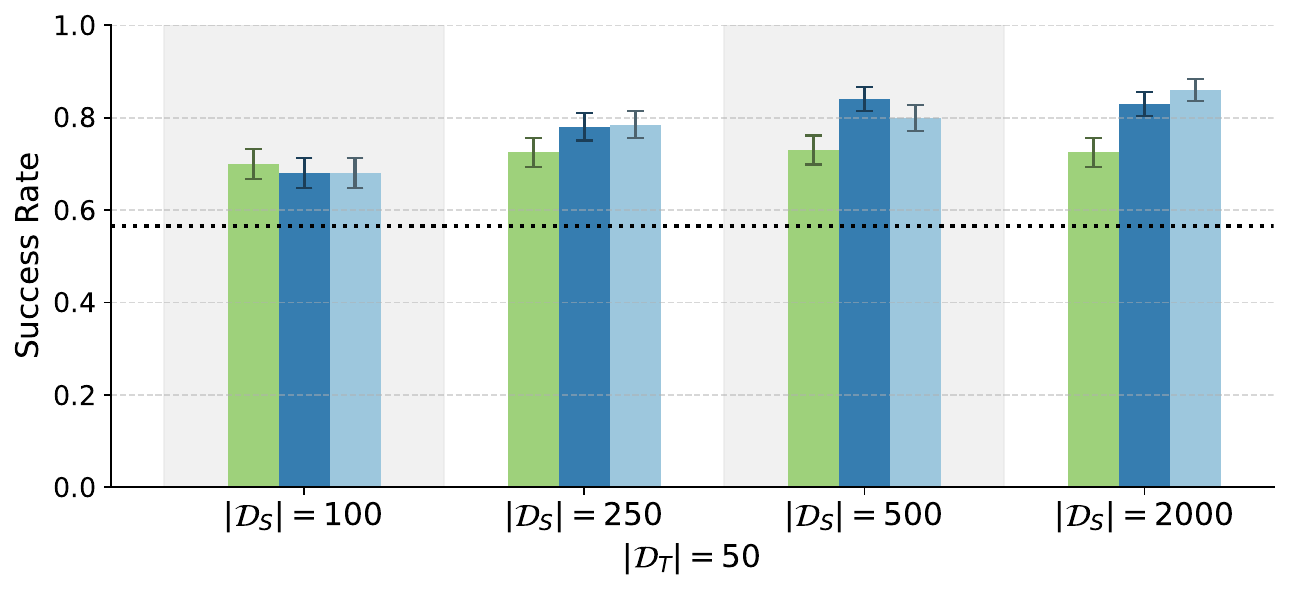}
    \caption{Comparing cotraining and finetuning in simulation. The results are consistent with the real-world experiments.\vspace{-\baselineskip}}
    \label{fig:scaled_finetuning}
\end{figure}

\begin{figure}[t]
    \centering
    \includegraphics[width=\linewidth]{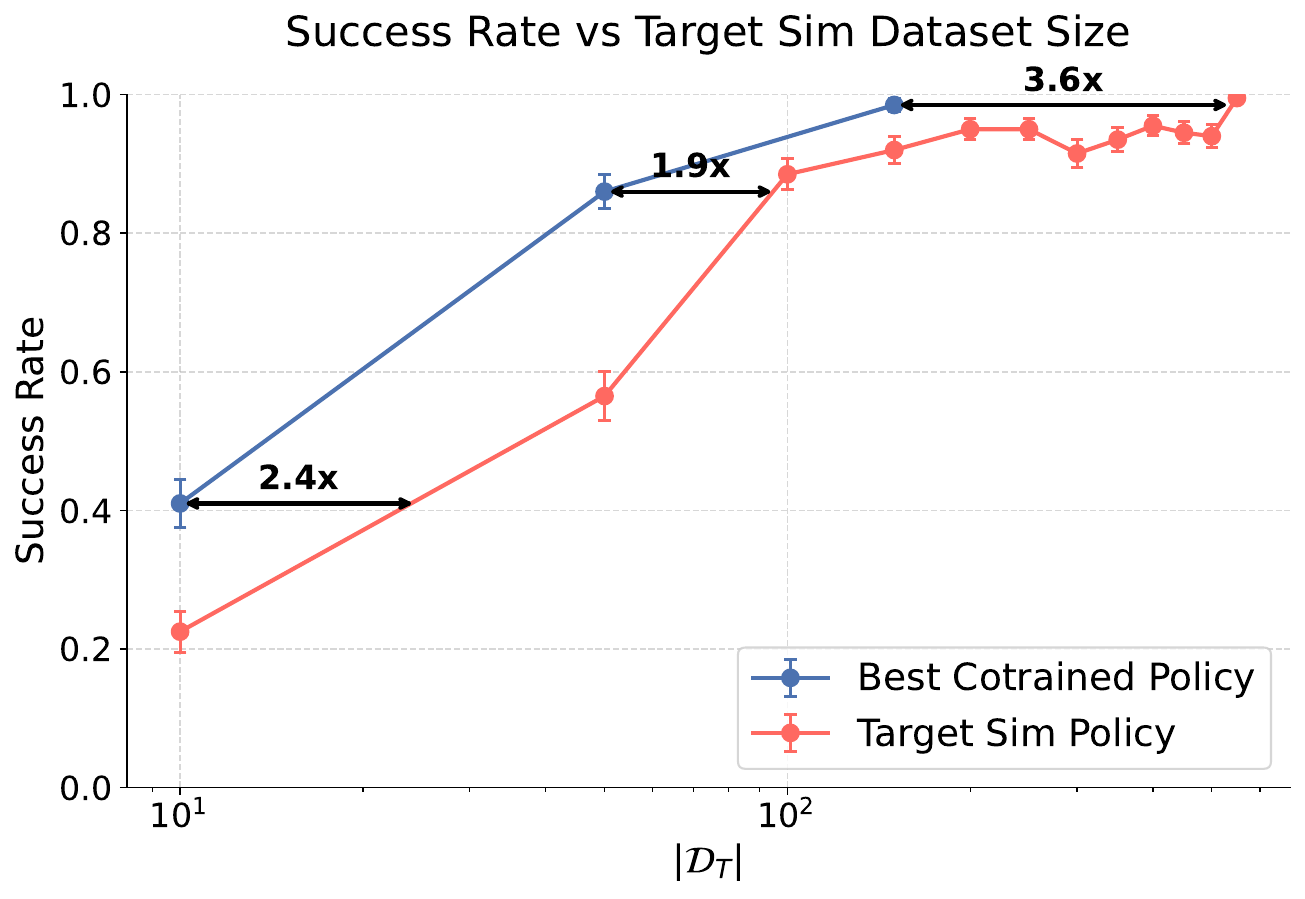}
    \caption{The performance of the best cotrained policies are equivalent to increasing the amount of data from the target domain by 2-3x.\vspace{-\baselineskip}}
    \label{fig:cotraining_plateaus}
\end{figure}

The results in Figure \ref{fig:scaled_cotraining} mostly agree with Section \ref{real_world:results}. We highlight key differences. First, cotraining increased performance even in the high-data regime; but the improvement in the other data regimes was lower compared to Section \ref{sec:real_world}. Second, the optimal $\alpha$'s were lower. We hypothesize that this is because the \textit{sim2target} gap was smaller than the \textit{sim2real} gap. The finetuning results in Figure \ref{fig:scaled_finetuning} are consistent with Section \ref{real_world:finetuning}.

Scaling $\mathcal D_S$ improves performance, but this trend eventually plateaus. This suggests that sim data cannot replace real data; real data is still needed to increase the cotraining ceiling. Figure \ref{fig:cotraining_plateaus} compares the performance plateaus at different data scales to the performance of policies trained on only \textit{target sim} data. In our specific setup, the benefits from cotraining with sim data were equivalent to increasing our \textit{target sim} dataset by roughly 2-3x.

\subsection{Distribution Shift Experiments}
\label{sim:distribution_shifts}
We use simulation to answer the following questions about the effects of distribution shifts on sim-and-real cotraining:
\begin{enumerate}
    \item Which sim2real gaps matter for cotraining and how do they inform simulator design for data generation?
    \item What are best practices for cotraining under different types and magnitudes of sim2real gaps?
\end{enumerate}

\begin{figure}[t]
    \centering
    \includegraphics[width=\linewidth]{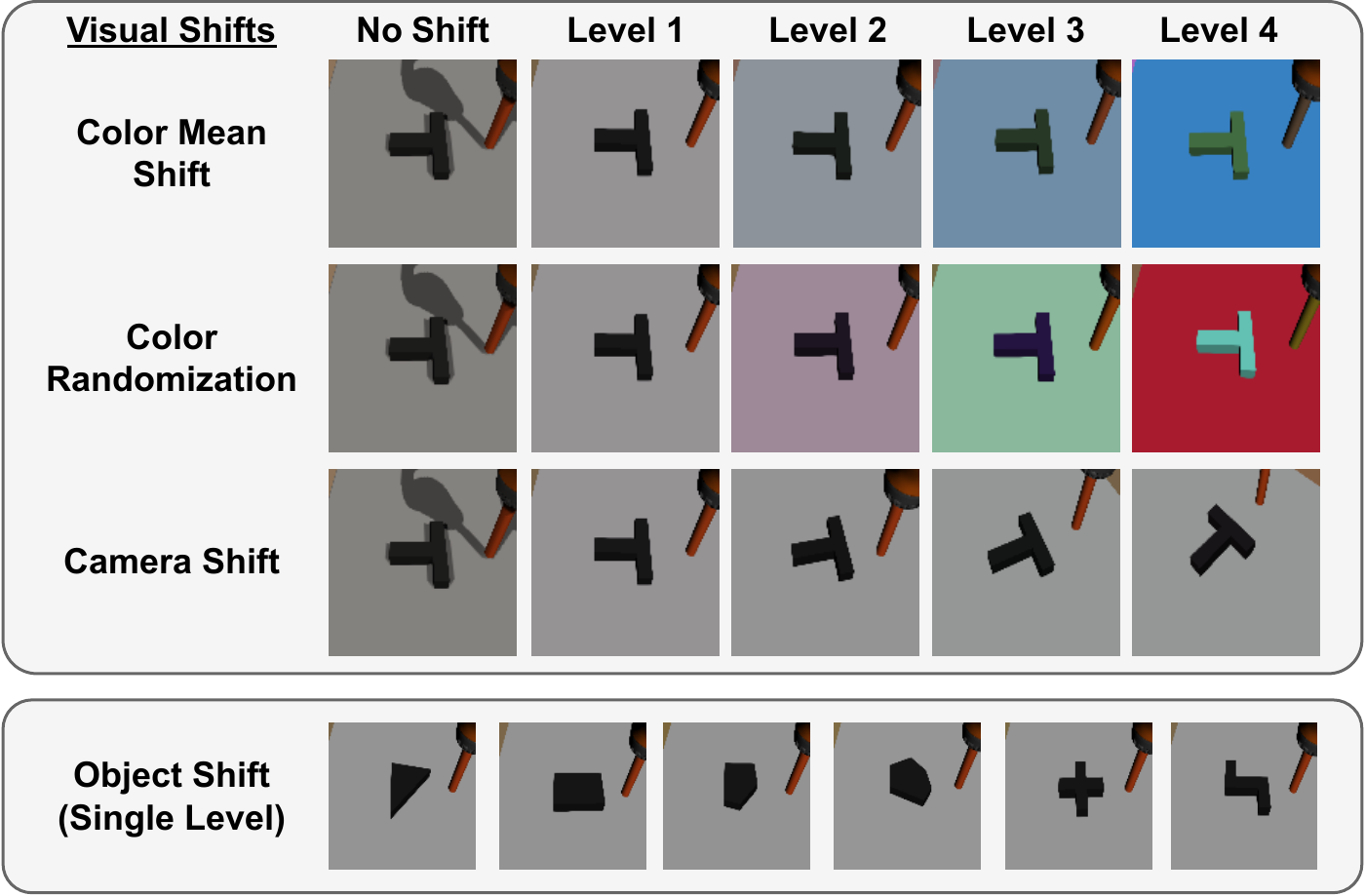}
    \caption{The upper box visualizes the 3 types of visual shifts over 4 intensity levels. The bottom box visualizes the sliders used for object shift.\vspace{-\baselineskip}}
    \label{fig:sim_sim_visual_shift}
\end{figure}

We explore 6 distribution shifts at varying intensities. These shifts were partially inspired by \cite{zhao2022representationlearningvideos}. Let $\mathcal C=[0,1]^3$ be the space of RGB colors, $B_r(\boldsymbol{c})$ be a ball of radius $r$ centered at $\boldsymbol{c}$, and $\boldsymbol{c}_i\in \mathcal C$ be the true color of object $i$.
\begin{itemize}
    \item \textbf{Color Mean Shift:} Object colors are shifted as $\boldsymbol{c}_i\leftarrow \boldsymbol{c}_i + \gamma\boldsymbol{u}_i + \boldsymbol{o}_i$, where $\boldsymbol{u}_i$'s are unit vectors. $\boldsymbol{o}_i=-0.05\cdot\boldsymbol{1}$ for the slider and $\boldsymbol{0}$ otherwise. We sweep $\gamma\in\{0, 0.05, 0.2, 0.5\}$.
    \item \textbf{Color Randomization:} Object colors are sampled uniformly from $B_r(\boldsymbol{c_i})\cap\mathcal C$. We sweep $r\in\{0.025, 0.1, 0.5, \sqrt 3\}$. Note that $B_{\sqrt3}(\boldsymbol{c_i})\cap\mathcal C = \mathcal C$.
    \item \textbf{Camera Shift:} We fix a frame, $F$, near the slider's target pose and rotate the camera about the $z$-axis of $F$ by $\beta\in\{0, -\tfrac{\pi}{16}, -\tfrac{\pi}{8}, -\tfrac{\pi}{4}\}$.
    \item \textbf{Center of Mass (CoM) Shift:} We generate data with incorrect CoM: $y_{\mathrm{CoM}} \leftarrow y_{\mathrm{CoM}}+y_{\mathrm{offset}}$. We sweep $y_{\mathrm{offset}}\in\{0, 3, -3, -6\}$cm. The slider is 16.5cm tall.
    \item \textbf{Goal Shift: }We generate demos that push the slider to the incorrect goal pose: $y_{\mathrm{goal}} \leftarrow y_{\mathrm{goal}}+g_{\mathrm{offset}}$. We sweep $g_{\mathrm{offset}}\in\{0, -2.5, -5, -10\}$cm.
    \item \textbf{Object Shift: }We generate a planar pushing dataset containing 6 objects, none of which are the T.
\end{itemize}
These 6 shifts fall under 3 categories of sim2real gap. Color mean, color randomization, and camera shift are examples of \textit{visual} gap; CoM shift is an example of a \textit{physics} gap; goal and object shift are examples of \textit{task} gap. We visualize 4 of the 6 shifts in Figure \ref{fig:sim_sim_visual_shift}.

\begin{figure}[t]
    \centering
    {\includegraphics[width=\linewidth]{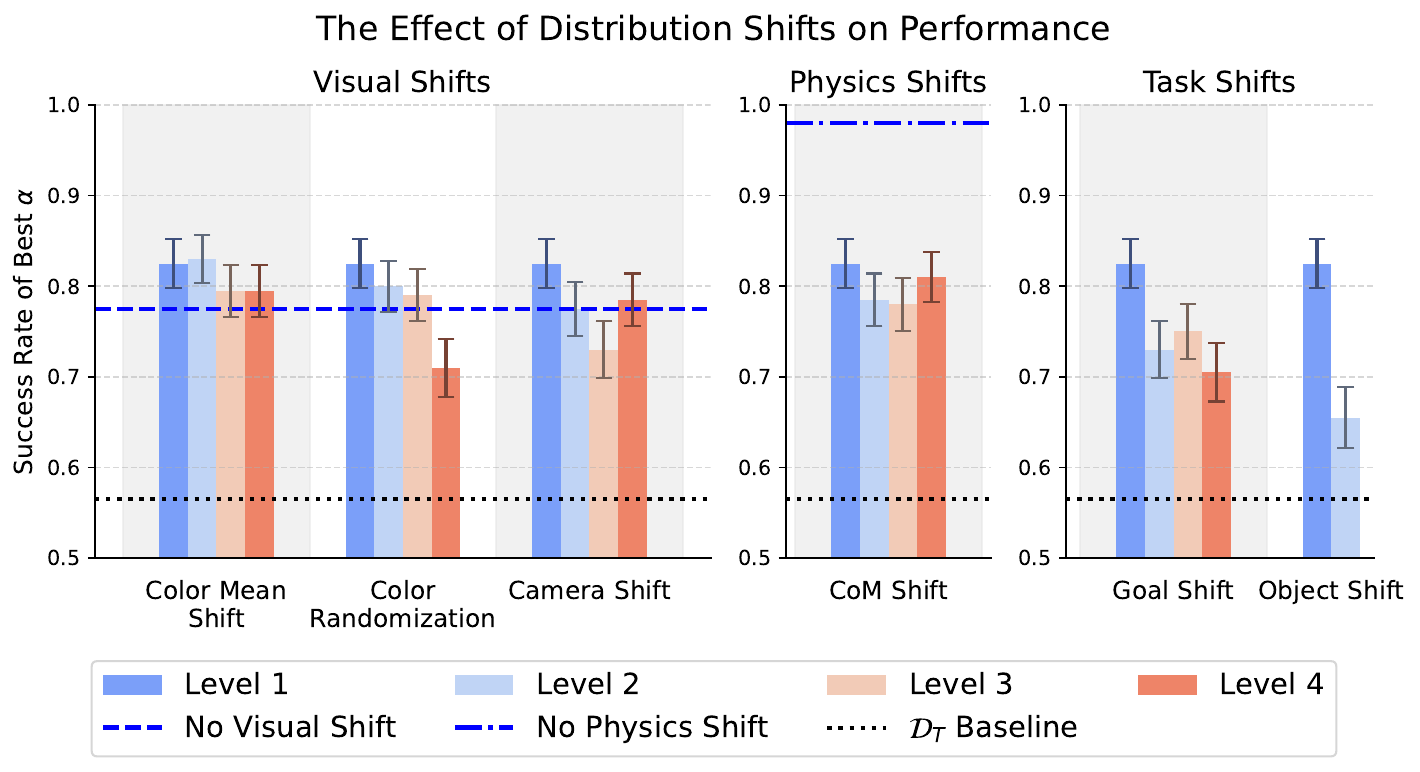}}
    \caption{Performance of the best data mixture for each distribution shift and intensity.}
    \label{fig:distribution_shifts_results}
\end{figure}

We collect a single \textit{target} dataset, $\mathcal D_T$, with 50 demos and reuse it for all policies in this experiment. Distribution shifts are introduced by modifying $\mathcal D_S$. Concretely, we set all shifts to their lowest values (Level 1) and sweep the remaining levels of each shift individually. Figure \ref{fig:distribution_shifts_results} reports the performance of the best data mixture for each shift. All policies were trained with $|\mathcal D_T|=50$ and $|\mathcal D_S| =2000$.

For Level 1 CoM shift, a physics gap remains since the sim and \textit{target} sim environment use different physics models (see Table \ref{tab:sim2sim_gap}). Similarly for Level 1 visual shifts, a visual gap remains since the \textit{target} environment contains shadows while the sim environment does not (see Figure \ref{fig:sim_sim_visual_shift}). To study the effect of completely eliminating visual or physics gaps, we collect 2 additional simulated datasets: one with \textit{no visual gap}, and one with \textit{no physics gap}. We cotrain policies on these datasets and present them in Figure \ref{fig:distribution_shifts_results} as well.

Overall, larger sim2real gaps reduce performance. This agrees with existing theoretical bounds \cite{theorydifferentdomains} and general intuition. We highlight a few noteworthy trends.

1) Performance is sensitive to task and physics shifts, and least sensitive to color mean shift. Optimal $\alpha$'s increased with color randomization and goal shift, but we did not observe strong trends for the other shifts. Unfortunately, calibrating the \emph{magnitude} of the visual shifts with the physical shifts is difficult since their units differ, but we have attempted to make both distributions reasonably representative.

2) The policy trained with no physics shift achieves 15.5\% higher success rate than the policy trained with Level 1 shift. This is a significant difference in success rate. Performance is less sensitive to subsequent increases in the physics gap. Nonetheless, the magnitude of the initial drop suggests that accurate simulation physics are important for contact-rich tasks. Tasks that emphasize semantic or visual reasoning and tasks that use prehensile or stabilizing grasps might exhibit less performance degradation.

3) Roughly speaking, performance improves with better rendering. This is consistent with findings from previous works \cite{qureshi2024splatsimzeroshotsim2realtransfer}\cite{rao2020rlcycleganreinforcementlearningaware}; but paradoxically, the policy cotrained with no visual gap is slightly weaker. This suggests that a small amount of visual gap is desirable. Without it, the policy cannot distinguish between the two environments. We hypothesize that this harms action prediction. This is a subtle, but important point that deserves more discussion. We examine it more closely in Section \ref{sec:analysis}.

Together, Section \ref{sim:distribution_shifts} and the upcoming discussions in Section \ref{analysis:sim-real} suggest that simulated environments for cotraining should match the physics as closely as possible, and provide enough information to distinguish sim-and-real. Improved rendering increases performance; however, \textit{perfect} rendering is unnecessary and difficult to achieve in practice \cite{qureshi2024splatsimzeroshotsim2realtransfer}. Policies remained sensitive to the mixing ratio; but, the optimal mixing ratio was largely unaffected by most shifts.

\section{How does cotraining succeed?}
\label{sec:analysis}
We analyze factors that contributed to the success of cotraining. Section \ref{analysis:sim-real} shows that a policy's ability to identify its domain is crucial. High-performing policies behave distinctly more like $\mathcal D_R$ when deployed in the real-world, and more like $\mathcal D_S$ when deployed in sim. These findings were also true in our simulated experiments. We remind the reader that $\mathcal D_S$ and $\mathcal D_R$ contain noticeable action gap.

If cotrained policies learn distinct behaviors for each domain, then how does training from one improve performance in the other? Section \ref{analysis:mechanisms} discusses two mechanisms that facilitate this positive transfer: 1) \textit{dataset coverage}, 2) \textit{power laws} on real-world performance metrics.

\subsection{Sim-and-Real Discernability}
\label{analysis:sim-real}
\begin{figure}[t]
    \centering
    \includegraphics[width=\linewidth]{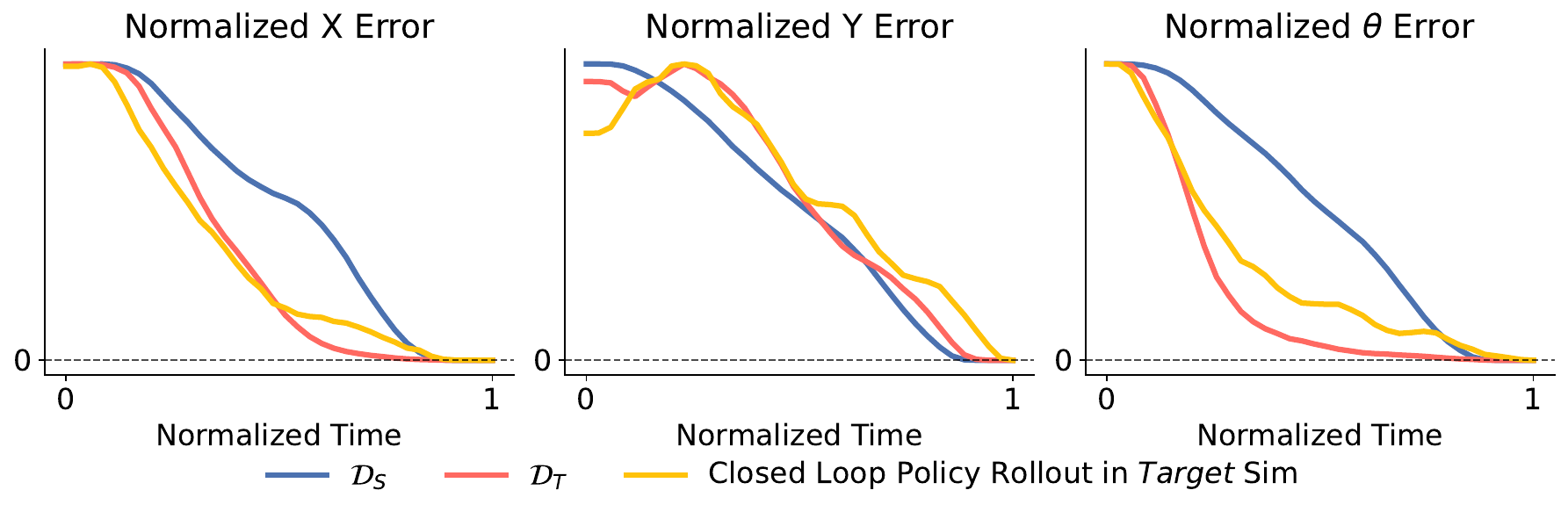}
    \caption{The red and blue lines show the average normalized slider error over time for demonstrations in the \textit{target} and simulated datasets respectively. The yellow line plots the same statistic for 200 rollouts in the \textit{target} sim environment for a cotrained policy with $|\mathcal D_T|=50$, $|\mathcal D_S|=2000$, $\alpha=0.024$. Although $|\mathcal{D_S}| \gg |\mathcal{D}_T|$ and $\alpha\approx0$, the yellow line is closer to the error for $\mathcal{D}_T$ (red) than $\mathcal{D_S}$ (blue).\vspace{-0.5\baselineskip}}
    \label{fig:absolute_error_plots}
\end{figure}

\begin{table}[t]
    \centering
    \begin{tabular}{|c|c|c|c|c|}
        \hline
        \multicolumn{3}{|c|}{\textbf{Policy}} & \multicolumn{2}{|c|}{\textbf{Binary Probe Location}} \\ \hline
        $|\mathcal D_R|$ & $|\mathcal D_S|$ & $\alpha$ & Observation Embedding & Final Activation \\ \hline
        50 & 500 & 0.75& 100\% & 74.2\% \\ \hline
        50 & 2000 & 0.75& 100\% & 89\% \\ \hline
        10 & 500 & 0.75& 100\% & 84\% \\ \hline
        10 & 2000 & 5e-3& 100\% & 93\% \\ \hline
    \end{tabular}
    \caption{The environment classification accuracy for binary probes at different layers of several cotrained policies.\vspace{-0.5\baselineskip}}
    \label{tab:binary_probing}
\end{table}

Figure \ref{fig:absolute_error_plots} demonstrates that the behaviors of cotrained policies in the \textit{target} sim environment are distinctly closer to $\mathcal D_T$ than $\mathcal D_S$. Table \ref{tab:binary_probing} shows this is not a coincidence: binary probes on the learned embeddings of cotrained policies can accurately classify sim and real. Unsurprisingly, the observation embedding carries information about the environment label; however, policies choose to preserve this information (which would otherwise be destroyed by the data processing inequality) until the final activation. In other words, cotrained policies learn that the environment label is important for action prediction.

\begin{figure}[t]
    \centering
    \includegraphics[width=\linewidth]{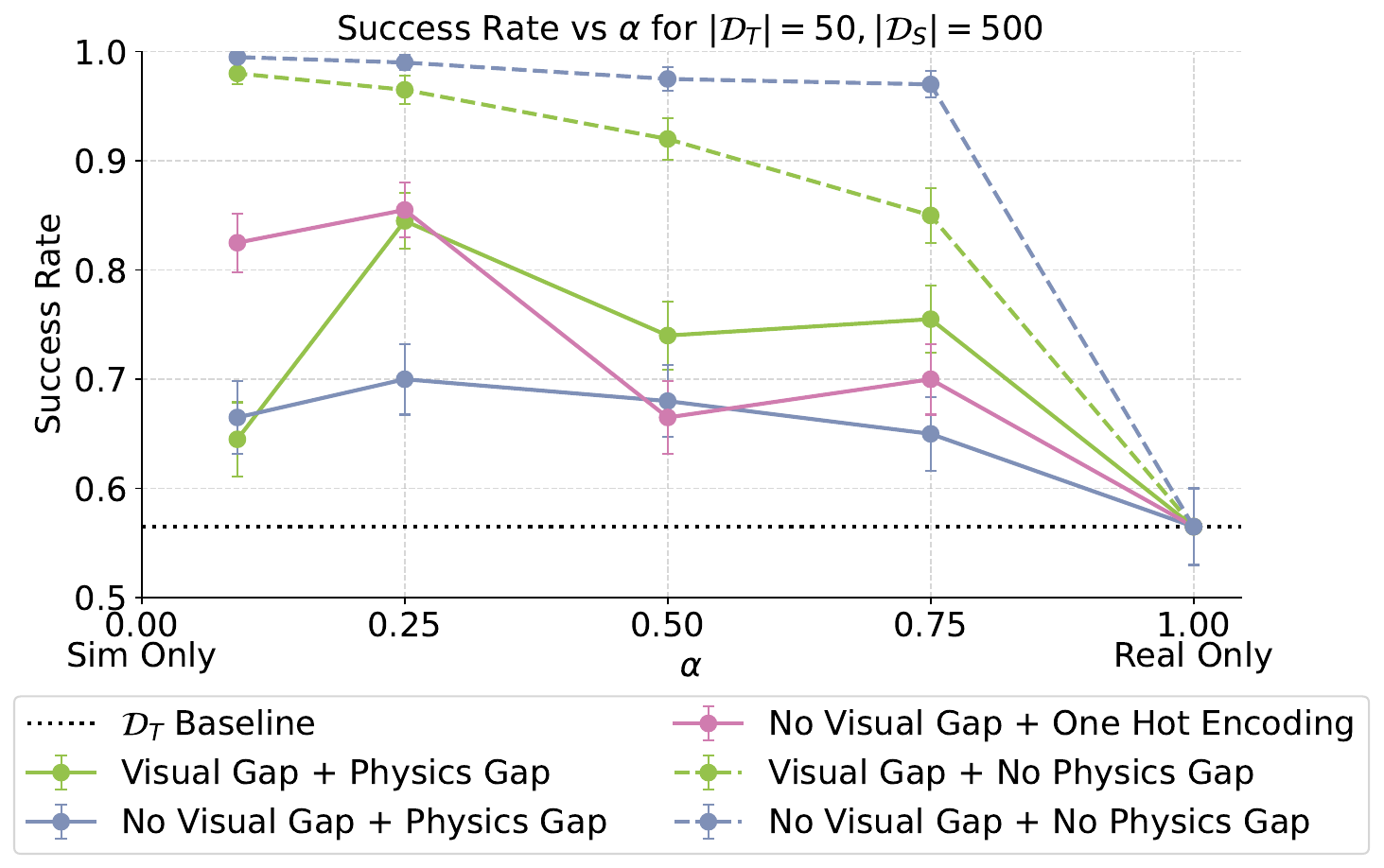}
    \caption{When a physics gap exists, removing visual gap harms performance. The opposite is true when a physics gap does not exist. This shows that visual gaps (or one-hot encodings) are important for enabling cotrained policies to identify the underlying physics of their deployment domain.\vspace{-1.5\baselineskip}}
    \label{fig:one_hot}
\end{figure}

High-performing policies must differentiate sim from real since each environment's physics require
different actions. Figure \ref{fig:one_hot} supports this intuition. We first discuss the case when $\mathcal D_S$ and $\mathcal D_T$ contain physics gap (solid lines in Figure \ref{fig:one_hot}). In this setting, removing visual differences between sim and \textit{target} decreases success rate. Adding a one-hot encoding for the environment recovers this performance. This shows that sim-and-real discernibility is important.

On the other hand, when $\mathcal D_S$ and $\mathcal D_T$ do not contain physics gap, the opposite trend emerges: removing the visual gap improves performance. This supports the second part of our claim: sim-and-real discernibility is important because the physics between the two environments are different.

\subsection{Mechanisms For Positive Transfer in Cotraining}
\label{analysis:mechanisms}
Although cotrained policies learn different actions for each domain, learning from sim still improves performance. We discuss mechanisms and evidence for this positive transfer.

\subsubsection{Coverage}
\begin{figure}[t]
    \centering
    \includegraphics[width=\linewidth,trim={0cm 0.25cm 0cm 0.25cm},clip]{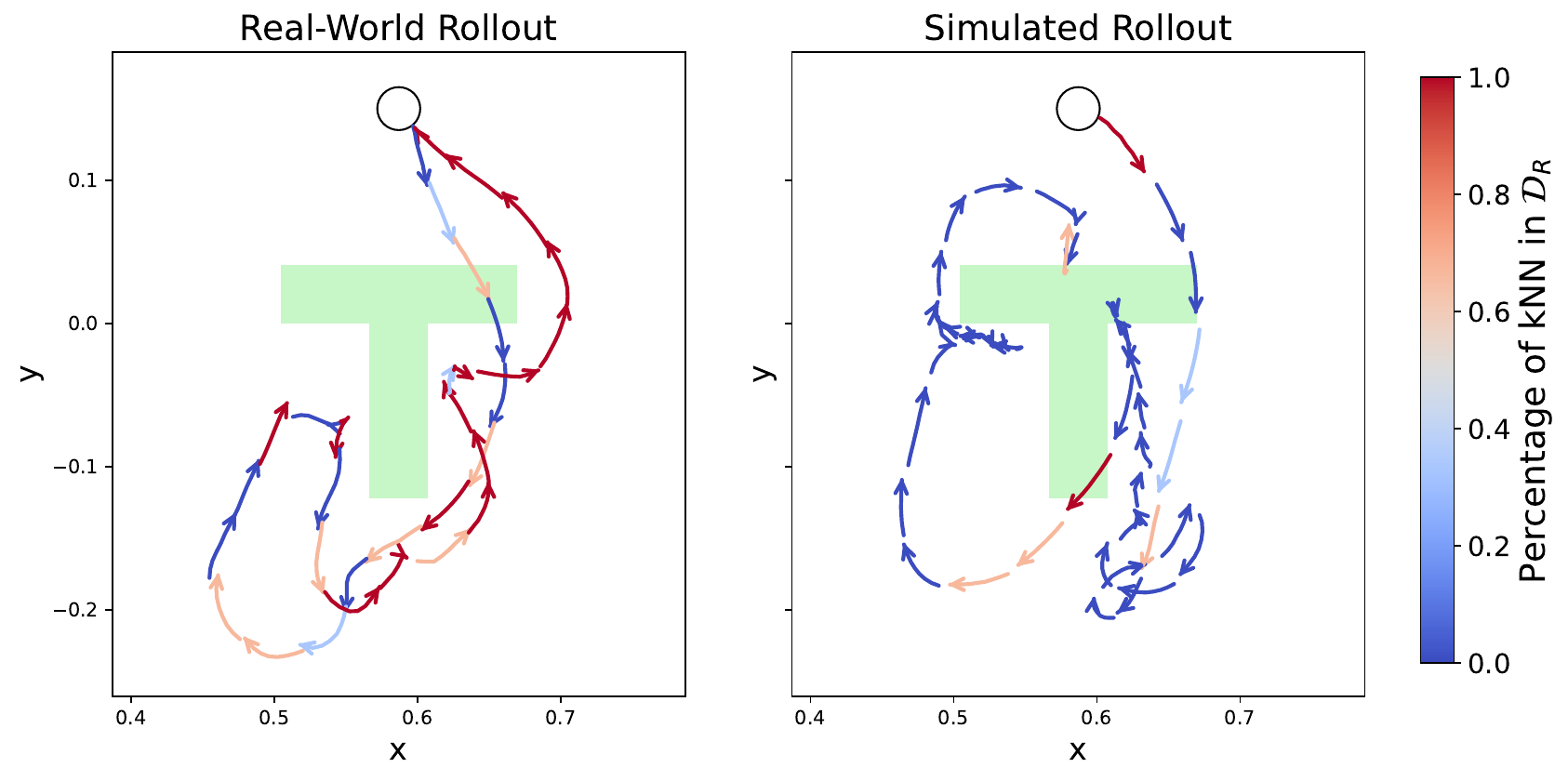}
    \caption{This figure visualizes when actions were learned from $\mathcal D_R$ (red) or $\mathcal D_S$ (blue). Each arrow is a chunk of predicted actions and its color indicates the percentage of its nearest neighbors in $\mathcal D_R$ vs $\mathcal D_S$. The green T is the goal pose and the circle is the robot's default position. The policy was trained with $|\mathcal D_R|=50$, $|\mathcal D_S|=500$, and $\alpha=0.75$.}\vspace{-\baselineskip}
    \label{fig:knn}
\end{figure}

We hypothesize that cotraining with $\mathcal D_S$ improves performance by filling gaps in $\mathcal D_R$. In other words, a policy deployed in real learns most of its behavior from $\mathcal D_R$ (see Figure \ref{fig:absolute_error_plots}) and relies on $\mathcal D_S$ in states that were not covered by $\mathcal D_R$. We illustrate this by analyzing when a policy's output was learned from real vs sim. We assume that predicting action $\mathbf{A}$ given observation $\mathbf{O}$ is learned from $\mathcal D_R$ if more of its nearest observation-action neighbors are in $\mathcal D_R$, and vice-versa for $\mathcal D_S$\footnote{Given $(\mathbf O_1, \mathbf A_1)$ and $(\mathbf O_2, \mathbf A_2)$, we compute their distance as $\lVert\mathbf A_1 - \mathbf A_2\rVert_2 + \tau\lVert g(\mathbf O_1)-g(\mathbf O_2)\rVert_2$, where $g$ is the observation embedding.}. We roll out a cotrained policy twice: once in the real-world and once in the simulation. Figure \ref{fig:knn} visualizes the percentage of the top-$k$ ($k=3)$ nearest-neighbors from $\mathcal D_R$ vs $D_S$ for every chunk of predicted actions. The visualizations for other values of $k$ are similar.

The real-world rollout is mostly red (learned from $\mathcal D_R$) and interleaved with blue (learned from $\mathcal D_S$). The opposite is true for the simulation rollout. This result is notable since the policy was cotrained with 10x more sim data than real data; yet the majority of the nearest neighbors from the real-world rollout were still from $\mathcal D_R$.

Figure \ref{fig:knn} suggests that policies primarily rely on data from their deployment domain. Cotraining improves performance by providing actions when the policy encounters missing states from the deployment domain's dataset. This helps explain why performance plateaus even as $|\mathcal D_S|$ grows: once missing states in $\mathcal D_R$ are filled, additional simulated data provides diminishing returns. It also explains why policies cotrained with $|\mathcal D_R|=10$ perform poorly: simulation can fill gaps, but we need real-world data to learn compatible strategies and behaviors for the real-world physics.

\subsubsection{Power Laws} Despite the \textit{sim2target} gap, scaling simulation predictably decreases the \textit{test loss} and the \textit{action mean squared error (MSE)} in the \textit{target} domain. This is evidence for positive transfer in cotraining. The test loss is the denoiser loss \eqref{eq:denoiser_loss} achieved on $\mathcal D_T^\mathrm{test}$. Similarly, action MSE is the error of fully denoised actions in $\mathcal D_T^\mathrm{test}$. Importantly, no checkpoints were trained or selected with the test dataset.

\begin{figure}[t]
    \centering
    \includegraphics[width=0.49\linewidth]{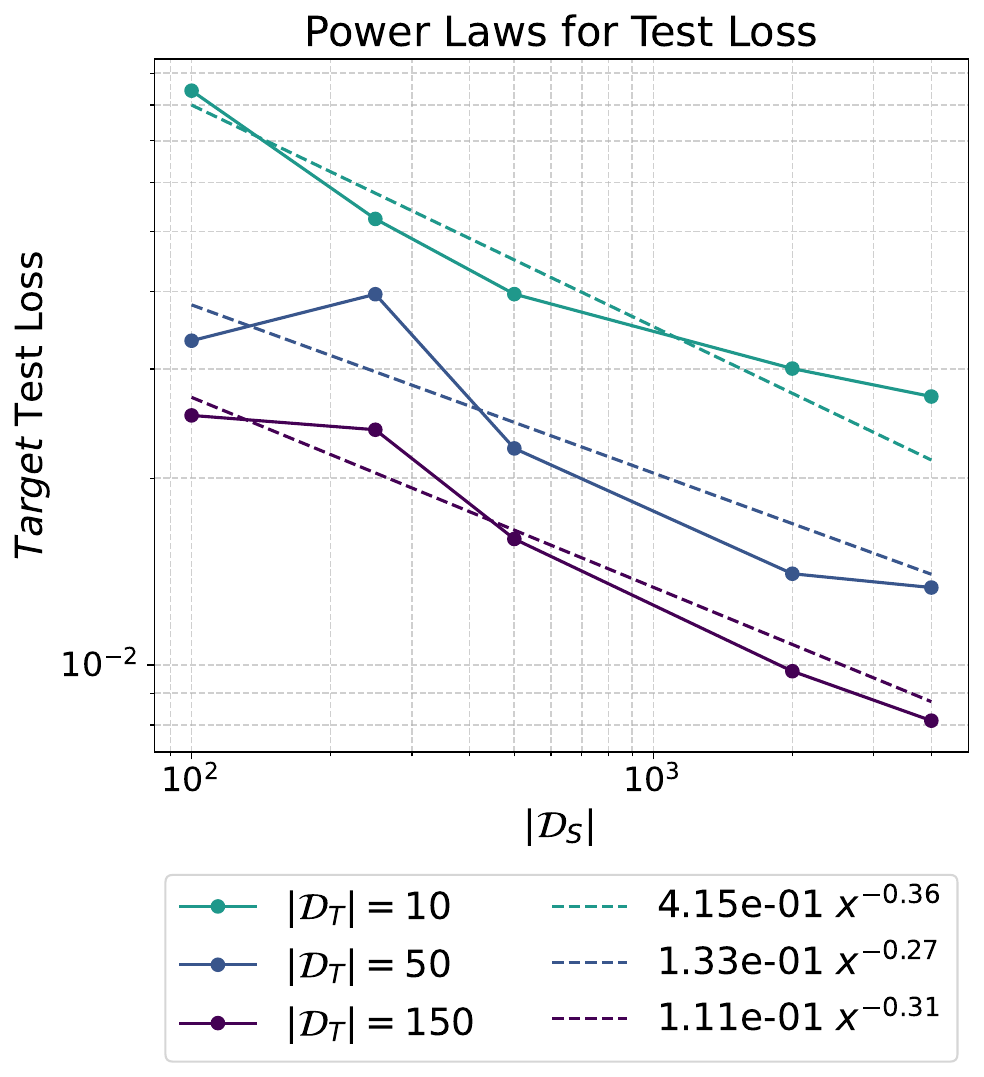}
    \includegraphics[width=0.49\linewidth]{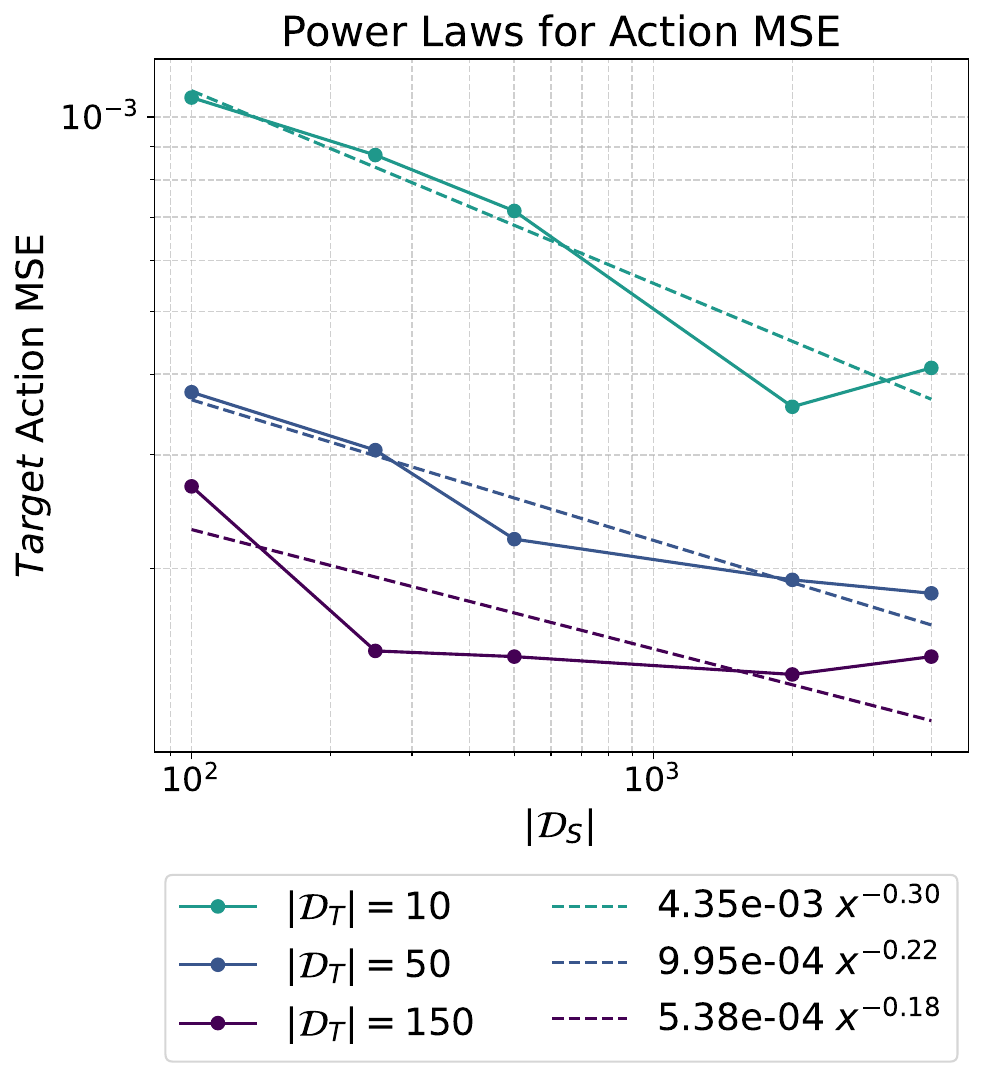}
    \caption{Log-log plots of the \textit{target} test loss and action MSE as a function of $|\mathcal D_S|$. We report values for all the checkpoints in Figure \ref{fig:scaled_cotraining} that were trained with the natural mixing ratio. Note that the plots exhibit an approximate power law.}\vspace{-\baselineskip}
    \label{fig:power_laws}
\end{figure}

Figure \ref{fig:power_laws} visualizes the power laws. Test loss and action MSE decrease predictably with similar exponents for each value of $|\mathcal D_T|$. Equation \eqref{eq:power_laws} shows the power laws w.r.t both $|\mathcal D_S|$ and $|\mathcal D_T|$. These equations provide intuition for the relative impact of scaling both datasets. For instance, the magnitude of the exponents on $|\mathcal D_T|$ are larger than $|\mathcal D_S|$. This aligns with our intuition that real-world data is more valuable than simulated data. We note that \eqref{eq:power_laws} is specific to our setup; nevertheless, the existence of power laws that accurately fit the performance is remarkable.
\begin{equation}
    \begin{aligned}
        \mathcal L_{\mathcal D_T^\mathrm{test}} \propto |\mathcal D_S|^{-0.332} \cdot |\mathcal D_T|^{-0.397}, \quad R^2 = 0.945 \\
        \mathrm{MSE}_{\mathcal D_T^\mathrm{test}} \propto |\mathcal D_S|^{-0.285} \cdot |\mathcal D_T|^{-0.587}, \quad R^2 = 0.975
    \end{aligned}
    \label{eq:power_laws}
\end{equation}

Larger experiments are needed to determine if these power laws are also \textit{scaling laws} \cite{kaplan2020scalinglawsneurallanguage}. For instance, $\mathrm{MSE}_{\mathcal D_T^\mathrm{test}}$ appears to plateau slightly for large $|\mathcal D_S|$. This could explain the performance plateaus in Section \ref{sim:cotraining_asymptotes}.

\section{Ablations}
\label{sec:ablations}

We explore 2 ablations inspired by our analysis: one that artifically \textit{amplifies} the sim2real gap (Section \ref{ablations:cfg}), and another that \textit{reduces} the sim2real gap at the representation level (Section \ref{ablations:alternative}).

\subsection{Classifier-Free Guidance Ablation}
\label{ablations:cfg}
If high-performing policies predict different actions in real vs sim, can we improve performance by amplifying this action gap with \textit{classifier-free guidance} (CFG) \cite{ho2022classifierfreediffusionguidance}? To study this question, we add a one-hot encoding of the environment to the policies and cotrain with CFG. More concretely, we sample with the denoiser in \eqref{eq:cfg_cotrain} which is \textit{guided} by the environment encoding, $c$.
\begin{equation}
    \tilde{\epsilon}(\mathbf A_t, \mathbf O, c, t) = (1+w)\epsilon_\theta(\mathbf A_t, \mathbf O, c, t) - w\epsilon_\theta(\mathbf A_t, \mathbf O, \varnothing, t).
\label{eq:cfg_cotrain}
\end{equation}

\begin{figure}[h]
    \centering\includegraphics[width=\linewidth]{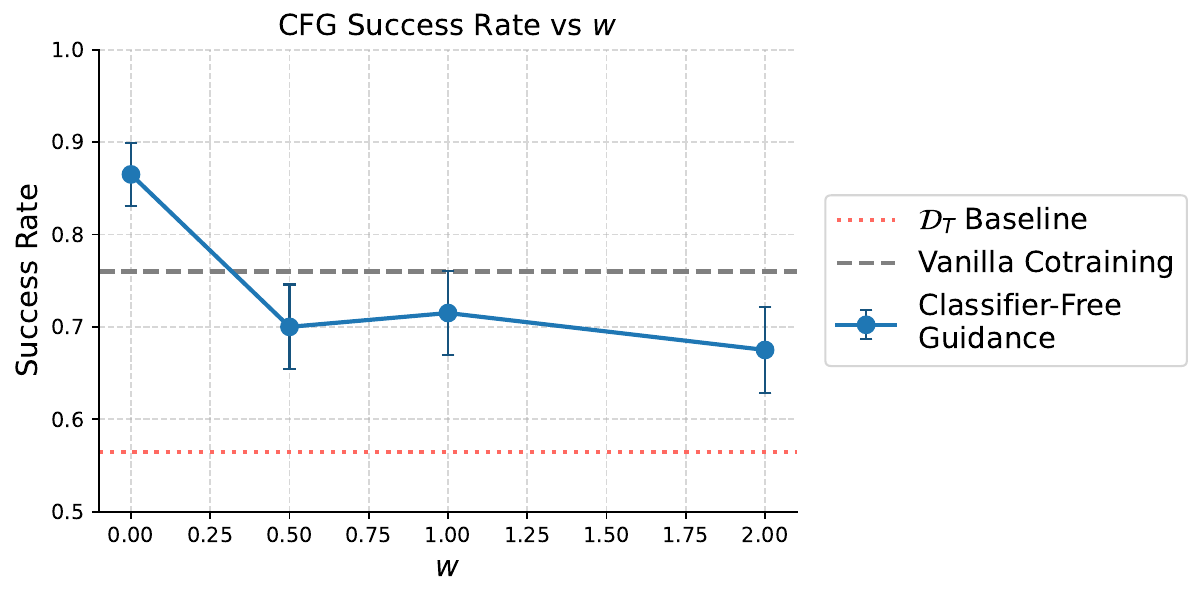}
    \caption{Performance of cotraining with CFG for $|\mathcal D_T|=50$, $|\mathcal D_S|=2000$, $\alpha=0.25$. The plots are similar for all data scales and mixtures. We present only one for clarity.\vspace{-0.5\baselineskip}}
    \label{fig:cfg}
\end{figure}

CFG artificially amplifies the action gap at sampling-time; $w$ controls the magnitude of this effect. Note that $w=0$ is equivalent to regular sampling with a one-hot encoding. Figure \ref{fig:cfg} shows that amplifying the action gap with CFG and $w>0$ does not improve performance over vanilla cotraining. On the other hand, preformance increases for $w=0$. This suggests that practitioners should cotrain policies with a \emph{one-hot encoding of the environment} and sample regularly \cite{chi2024diffusionpolicy}.

\subsection{Alternative Cotraining Loss}
\label{ablations:alternative}
We showed that smaller sim2real gaps improve performance, but high-performing policies must distinguish sim from real. Given these observations, we explored an alternative loss for cotraining that \textit{reduces} the sim2real gap at the policy's ``representation level'' \cite{edwards2016censoringrepresentationsadversary}, while maintaining its ability to identify relevant domain-specific information.

More concretely, given a cotrained policy $\pi_\theta$, let $p^S_\theta$ and $p^R_\theta$ be distributions over the learned embeddings for sim and real respectively. Our alternative loss adds a second term to the training objective, $\mathrm{dist}(p^S_\theta, p^R_\theta)$, that penalizes the differences in the learned representations for sim and real.
\begin{equation}
    \min_\theta \mathcal L_{\mathcal D^\alpha}(\theta) + \lambda\cdot\mathrm{dist}(p^S_\theta, p^R_\theta).
\label{eq:alternative_loss}
\end{equation}

The second term in \eqref{eq:alternative_loss} encourages the policy to ignore sim2real gaps that are irrelevant for control (ex. differences in lighting and texture). The first term encourages the policy to retain any information that is important for minimizing $\mathcal L_{\mathcal D^\alpha}$ (ex. the physics of the domain). The two competing objectives are weighted by $\lambda$.

We explore two choices for the $\mathrm{dist}$ function. Here, $\mathcal L_\mathrm{BCE}$ is the binary cross entropy loss:
\begin{equation}
    \begin{aligned}
        \mathrm{Adversarial:\ }&\mathrm{dist}(p^S_\theta, p^R_\theta) = \max_\phi- \mathcal{L}_\mathrm{BCE}(\phi, \theta)\\
        \mathrm{MMD:\ }& \mathrm{dist}(p^S_\theta, p^R_\theta) = \mathrm{MMD}(p^S_\theta, p^R_\theta)
    \end{aligned}
    \label{eq:gan_mmd}
\end{equation}

The adversarial formulation aims to learn a representation for both sim and real that cannot be reliably distinguished by a classifier, $d_\phi$. This objective is minimized with a GAN-like training algorithm \cite{edwards2016censoringrepresentationsadversary}\cite{goodfellow2014generativeadversarialnetworks}. The MMD formulation aims to minimize the \textit{maximum mean discrepancy} (MMD) between the learned representations \cite{JMLR:v13:gretton12a}. We use a mixutre of Gaussian kernels for the MMD kernel and minimize \eqref{eq:alternative_loss} via differentiation. We swept the $\lambda$ hyperparameter for the MMD loss, but not the adversarial loss since the latter encountered the typical GAN-like training instabilities.

\begin{figure}[t]
    \centering
    \includegraphics[width=0.49\linewidth]{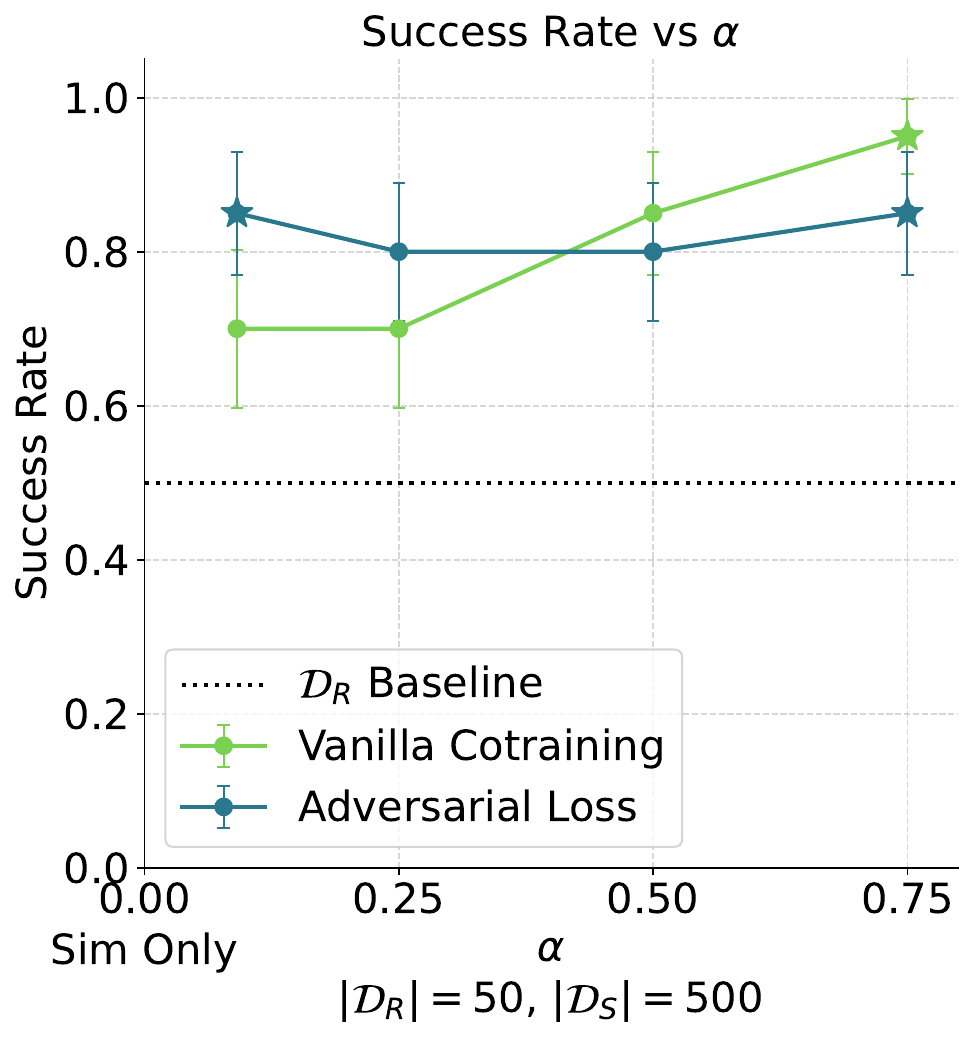}
    \includegraphics[width=0.49\linewidth]{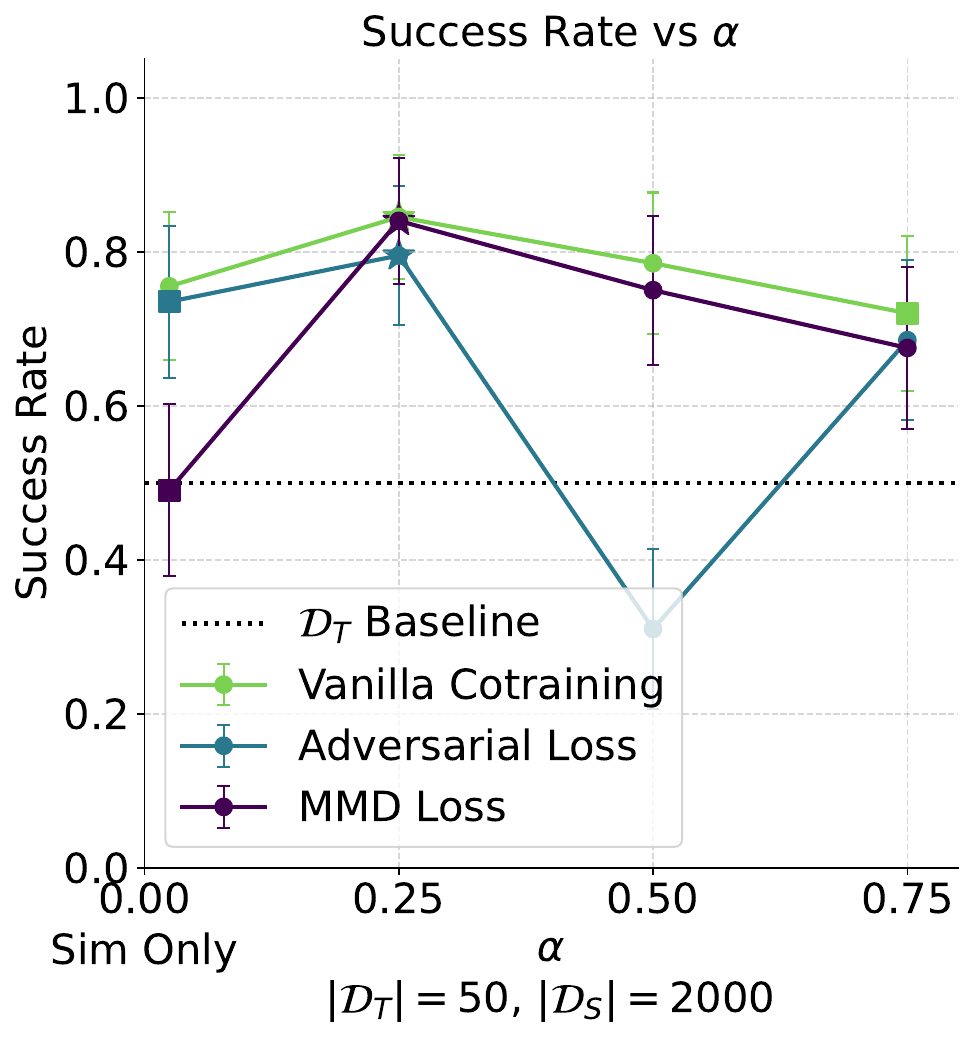}
    \caption{Performance of cotrained policies trained using the vanilla cotraining loss, adversarial loss, and MMD loss. Adding an MMD or GAN-like adversarial objective did not reliably improve performance in our experiments.}\vspace{-\baselineskip}
    \label{fig:alternative_loss}
\end{figure}

The performance of policies trained with \eqref{eq:alternative_loss} on both our real-world and \textit{target sim} environments are presented in Figure \ref{fig:alternative_loss}. The alternative loss formulations do not reliably outperform vanilla cotraining. This is consistent with our observation that sim-and-real discernability is important for performance.

Nonetheless, we chose to present these negative results for two reasons. First, the difference in performance between the adversarial and MMD formulations show that the choice of the $\mathrm{dist}$ function can dramatically alter the results. Thus, there could be unexplored choices for $\mathrm{dist}$ that outperform the vanilla cotraining baseline. Second, we believe the more general idea of reducing the sim2real gap at the representation level (as opposed to the simulation level) is an interesting direction for future work.

\section{Limitations}
\label{sec:limitations}
All evaluations were conducted on planar-pushing from pixels. This allows us to be thorough, but limits the scope of our results. Nonetheless, we hope our findings are informative for cotraining. A larger-scale evaluation would require a massively multi-task data generation pipeline. This is an exciting direction for future work. Repeating the experiments with other imitation learning algorithms \cite{lee2024behaviorgenerationlatentactions}\cite{zhao2023learningfinegrainedbimanualmanipulation} and data generation pipelines \cite{dalal2023imitatingtaskmotionplanning}\cite{ha2023scalingdistillingdownlanguageguided} would also be valuable.

\section{Conclusion}
\label{sec:conclusion}
We present a thorough empirical study of sim-and-real cotraining for robot imitation learning. Our results show that scaling up simulated data generation and cotraining are effective strategies for improving policy performance. We investigate the effects of mixing ratios, data scales, and distribution shifts. Lastly, we analyze the mechanisms underlying cotraining and present results for 2 alternative formulations for cotraining. We hope this work offers valuable insights for sim-and-real cotraining in robot learning.

\bibliographystyle{IEEEtran}
\bibliography{ref_with_url.bib}

\begin{thebibliography}{10}
\providecommand{\url}[1]{#1}
\csname url@samestyle\endcsname
\providecommand{\newblock}{\relax}
\providecommand{\bibinfo}[2]{#2}
\providecommand{\BIBentrySTDinterwordspacing}{\spaceskip=0pt\relax}
\providecommand{\BIBentryALTinterwordstretchfactor}{4}
\providecommand{\BIBentryALTinterwordspacing}{\spaceskip=\fontdimen2\font plus
\BIBentryALTinterwordstretchfactor\fontdimen3\font minus \fontdimen4\font\relax}
\providecommand{\BIBforeignlanguage}[2]{{%
\expandafter\ifx\csname l@#1\endcsname\relax
\typeout{** WARNING: IEEEtran.bst: No hyphenation pattern has been}%
\typeout{** loaded for the language `#1'. Using the pattern for}%
\typeout{** the default language instead.}%
\else
\language=\csname l@#1\endcsname
\fi
#2}}
\providecommand{\BIBdecl}{\relax}
\BIBdecl

\bibitem{reuther2018interactive}
A.~Reuther, J.~Kepner, C.~Byun, S.~Samsi, W.~Arcand, D.~Bestor, B.~Bergeron, V.~Gadepally, M.~Houle, M.~Hubbell, M.~Jones, A.~Klein, L.~Milechin, J.~Mullen, A.~Prout, A.~Rosa, C.~Yee, and P.~Michaleas, ``Interactive supercomputing on 40,000 cores for machine learning and data analysis,'' in \emph{2018 IEEE High Performance extreme Computing Conference (HPEC)}.\hskip 1em plus 0.5em minus 0.4em\relax IEEE, 2018, p. 1–6.

\bibitem{geminiteam2024geminifamilyhighlycapable}
\BIBentryALTinterwordspacing
G.~Team, ``Gemini: A family of highly capable multimodal models,'' 2024. [Online]. Available: \url{https://arxiv.org/abs/2312.11805}
\BIBentrySTDinterwordspacing

\bibitem{brown2020languagemodelsfewshotlearners}
\BIBentryALTinterwordspacing
T.~B. Brown, B.~Mann, N.~Ryder, M.~Subbiah, J.~Kaplan, P.~Dhariwal, A.~Neelakantan, P.~Shyam, G.~Sastry, A.~Askell, S.~Agarwal, A.~Herbert-Voss, G.~Krueger, T.~Henighan, R.~Child, A.~Ramesh, D.~M. Ziegler, J.~Wu, C.~Winter, C.~Hesse, M.~Chen, E.~Sigler, M.~Litwin, S.~Gray, B.~Chess, J.~Clark, C.~Berner, S.~McCandlish, A.~Radford, I.~Sutskever, and D.~Amodei, ``Language models are few-shot learners,'' 2020. [Online]. Available: \url{https://arxiv.org/abs/2005.14165}
\BIBentrySTDinterwordspacing

\bibitem{sun2017revisitingunreasonableeffectivenessdata}
\BIBentryALTinterwordspacing
C.~Sun, A.~Shrivastava, S.~Singh, and A.~Gupta, ``Revisiting unreasonable effectiveness of data in deep learning era,'' 2017. [Online]. Available: \url{https://arxiv.org/abs/1707.02968}
\BIBentrySTDinterwordspacing

\bibitem{firoozi2023foundationmodelsroboticsapplications}
\BIBentryALTinterwordspacing
R.~Firoozi, J.~Tucker, S.~Tian, A.~Majumdar, J.~Sun, W.~Liu, Y.~Zhu, S.~Song, A.~Kapoor, K.~Hausman, B.~Ichter, D.~Driess, J.~Wu, C.~Lu, and M.~Schwager, ``Foundation models in robotics: Applications, challenges, and the future,'' 2023. [Online]. Available: \url{https://arxiv.org/abs/2312.07843}
\BIBentrySTDinterwordspacing

\bibitem{dalal2023imitatingtaskmotionplanning}
\BIBentryALTinterwordspacing
M.~Dalal, A.~Mandlekar, C.~Garrett, A.~Handa, R.~Salakhutdinov, and D.~Fox, ``Imitating task and motion planning with visuomotor transformers,'' 2023. [Online]. Available: \url{https://arxiv.org/abs/2305.16309}
\BIBentrySTDinterwordspacing

\bibitem{ha2023scalingdistillingdownlanguageguided}
\BIBentryALTinterwordspacing
H.~Ha, P.~Florence, and S.~Song, ``Scaling up and distilling down: Language-guided robot skill acquisition,'' 2023. [Online]. Available: \url{https://arxiv.org/abs/2307.14535}
\BIBentrySTDinterwordspacing

\bibitem{mandlekar2023mimicgendatagenerationscalable}
\BIBentryALTinterwordspacing
A.~Mandlekar, S.~Nasiriany, B.~Wen, I.~Akinola, Y.~Narang, L.~Fan, Y.~Zhu, and D.~Fox, ``Mimicgen: A data generation system for scalable robot learning using human demonstrations,'' 2023. [Online]. Available: \url{https://arxiv.org/abs/2310.17596}
\BIBentrySTDinterwordspacing

\bibitem{embodimentcollaboration2024openxembodimentroboticlearning}
\BIBentryALTinterwordspacing
E.~Collaboration, A.~O'Neill, A.~Rehman, A.~Gupta, A.~Maddukuri, A.~Gupta, A.~Padalkar, A.~Lee, A.~Pooley, A.~Gupta, A.~Mandlekar, A.~Jain, A.~Tung, A.~Bewley, A.~Herzog, A.~Irpan, A.~Khazatsky, A.~Rai, A.~Gupta, A.~Wang, A.~Kolobov, A.~Singh, A.~Garg, A.~Kembhavi, A.~Xie, A.~Brohan, A.~Raffin, A.~Sharma, A.~Yavary, A.~Jain, A.~Balakrishna, A.~Wahid, B.~Burgess-Limerick, B.~Kim, B.~Schölkopf, B.~Wulfe, B.~Ichter, C.~Lu, C.~Xu, C.~Le, C.~Finn, C.~Wang, C.~Xu, C.~Chi, C.~Huang, C.~Chan, C.~Agia, C.~Pan, C.~Fu, C.~Devin, D.~Xu, D.~Morton, D.~Driess, D.~Chen, D.~Pathak, D.~Shah, D.~Büchler, D.~Jayaraman, D.~Kalashnikov, D.~Sadigh, E.~Johns, E.~Foster, F.~Liu, F.~Ceola, F.~Xia, F.~Zhao, F.~V. Frujeri, F.~Stulp, G.~Zhou, G.~S. Sukhatme, G.~Salhotra, G.~Yan, G.~Feng, G.~Schiavi, G.~Berseth, G.~Kahn, G.~Yang, G.~Wang, H.~Su, H.-S. Fang, H.~Shi, H.~Bao, H.~B. Amor, H.~I. Christensen, H.~Furuta, H.~Bharadhwaj, H.~Walke, H.~Fang, H.~Ha, I.~Mordatch, I.~Radosavovic, I.~Leal, J.~Liang, J.~Abou-Chakra, J.~Kim, J.~Drake,
  J.~Peters, J.~Schneider, J.~Hsu, J.~Vakil, J.~Bohg, J.~Bingham, J.~Wu, J.~Gao, J.~Hu, J.~Wu, J.~Wu, J.~Sun, J.~Luo, J.~Gu, J.~Tan, J.~Oh, J.~Wu, J.~Lu, J.~Yang, J.~Malik, J.~Silvério, J.~Hejna, J.~Booher, J.~Tompson, J.~Yang, J.~Salvador, J.~J. Lim, J.~Han, K.~Wang, K.~Rao, K.~Pertsch, K.~Hausman, K.~Go, K.~Gopalakrishnan, K.~Goldberg, K.~Byrne, K.~Oslund, K.~Kawaharazuka, K.~Black, K.~Lin, K.~Zhang, K.~Ehsani, K.~Lekkala, K.~Ellis, K.~Rana, K.~Srinivasan, K.~Fang, K.~P. Singh, K.-H. Zeng, K.~Hatch, K.~Hsu, L.~Itti, L.~Y. Chen, L.~Pinto, L.~Fei-Fei, L.~Tan, L.~J. Fan, L.~Ott, L.~Lee, L.~Weihs, M.~Chen, M.~Lepert, M.~Memmel, M.~Tomizuka, M.~Itkina, M.~G. Castro, M.~Spero, M.~Du, M.~Ahn, M.~C. Yip, M.~Zhang, M.~Ding, M.~Heo, M.~K. Srirama, M.~Sharma, M.~J. Kim, N.~Kanazawa, N.~Hansen, N.~Heess, N.~J. Joshi, N.~Suenderhauf, N.~Liu, N.~D. Palo, N.~M.~M. Shafiullah, O.~Mees, O.~Kroemer, O.~Bastani, P.~R. Sanketi, P.~T. Miller, P.~Yin, P.~Wohlhart, P.~Xu, P.~D. Fagan, P.~Mitrano, P.~Sermanet, P.~Abbeel,
  P.~Sundaresan, Q.~Chen, Q.~Vuong, R.~Rafailov, R.~Tian, R.~Doshi, R.~Mart'in-Mart'in, R.~Baijal, R.~Scalise, R.~Hendrix, R.~Lin, R.~Qian, R.~Zhang, R.~Mendonca, R.~Shah, R.~Hoque, R.~Julian, S.~Bustamante, S.~Kirmani, S.~Levine, S.~Lin, S.~Moore, S.~Bahl, S.~Dass, S.~Sonawani, S.~Tulsiani, S.~Song, S.~Xu, S.~Haldar, S.~Karamcheti, S.~Adebola, S.~Guist, S.~Nasiriany, S.~Schaal, S.~Welker, S.~Tian, S.~Ramamoorthy, S.~Dasari, S.~Belkhale, S.~Park, S.~Nair, S.~Mirchandani, T.~Osa, T.~Gupta, T.~Harada, T.~Matsushima, T.~Xiao, T.~Kollar, T.~Yu, T.~Ding, T.~Davchev, T.~Z. Zhao, T.~Armstrong, T.~Darrell, T.~Chung, V.~Jain, V.~Kumar, V.~Vanhoucke, W.~Zhan, W.~Zhou, W.~Burgard, X.~Chen, X.~Chen, X.~Wang, X.~Zhu, X.~Geng, X.~Liu, X.~Liangwei, X.~Li, Y.~Pang, Y.~Lu, Y.~J. Ma, Y.~Kim, Y.~Chebotar, Y.~Zhou, Y.~Zhu, Y.~Wu, Y.~Xu, Y.~Wang, Y.~Bisk, Y.~Dou, Y.~Cho, Y.~Lee, Y.~Cui, Y.~Cao, Y.-H. Wu, Y.~Tang, Y.~Zhu, Y.~Zhang, Y.~Jiang, Y.~Li, Y.~Li, Y.~Iwasawa, Y.~Matsuo, Z.~Ma, Z.~Xu, Z.~J. Cui, Z.~Zhang, Z.~Fu, and Z.~Lin,
  ``Open x-embodiment: Robotic learning datasets and rt-x models,'' 2024. [Online]. Available: \url{https://arxiv.org/abs/2310.08864}
\BIBentrySTDinterwordspacing

\bibitem{khazatsky2024droidlargescaleinthewildrobot}
\BIBentryALTinterwordspacing
D.~Collaboration, ``Droid: A large-scale in-the-wild robot manipulation dataset,'' 2024. [Online]. Available: \url{https://arxiv.org/abs/2403.12945}
\BIBentrySTDinterwordspacing

\bibitem{openai2019solvingrubikscuberobot}
\BIBentryALTinterwordspacing
OpenAI and et. al, ``Solving rubik's cube with a robot hand,'' 2019. [Online]. Available: \url{https://arxiv.org/abs/1910.07113}
\BIBentrySTDinterwordspacing

\bibitem{rudin2022learningwalkminutesusing}
\BIBentryALTinterwordspacing
N.~Rudin, D.~Hoeller, P.~Reist, and M.~Hutter, ``Learning to walk in minutes using massively parallel deep reinforcement learning,'' 2022. [Online]. Available: \url{https://arxiv.org/abs/2109.11978}
\BIBentrySTDinterwordspacing

\bibitem{ankile2024imitationrefinementresidual}
\BIBentryALTinterwordspacing
L.~Ankile, A.~Simeonov, I.~Shenfeld, M.~Torne, and P.~Agrawal, ``From imitation to refinement -- residual rl for precise assembly,'' 2024. [Online]. Available: \url{https://arxiv.org/abs/2407.16677}
\BIBentrySTDinterwordspacing

\bibitem{chi2024diffusionpolicy}
C.~Chi, Z.~Xu, S.~Feng, E.~Cousineau, Y.~Du, B.~Burchfiel, R.~Tedrake, and S.~Song, ``Diffusion policy: Visuomotor policy learning via action diffusion,'' \emph{The International Journal of Robotics Research}, 2024.

\bibitem{ho2022classifierfreediffusionguidance}
\BIBentryALTinterwordspacing
J.~Ho and T.~Salimans, ``Classifier-free diffusion guidance,'' 2022. [Online]. Available: \url{https://arxiv.org/abs/2207.12598}
\BIBentrySTDinterwordspacing

\bibitem{zhu2024learncontactrichmanipulationpolicies}
\BIBentryALTinterwordspacing
H.~Zhu, T.~Zhao, X.~Ni, J.~Wang, K.~Fang, L.~Righetti, and T.~Pang, ``Should we learn contact-rich manipulation policies from sampling-based planners?'' 2024. [Online]. Available: \url{https://arxiv.org/abs/2412.09743}
\BIBentrySTDinterwordspacing

\bibitem{yang2025physics}
L.~Yang, H.~T. Suh, T.~Zhao, B.~P. Græsdal, T.~Kelestemur, J.~Wang, T.~Pang, and R.~Tedrake, ``Physics-driven data generation for contact-rich manipulation via trajectory optimization,'' \emph{arXiv preprint arXiv:2206.10787}, 2025.

\bibitem{torne2024reconcilingrealitysimulationrealtosimtoreal}
\BIBentryALTinterwordspacing
M.~Torne, A.~Simeonov, Z.~Li, A.~Chan, T.~Chen, A.~Gupta, and P.~Agrawal, ``Reconciling reality through simulation: A real-to-sim-to-real approach for robust manipulation,'' 2024. [Online]. Available: \url{https://arxiv.org/abs/2403.03949}
\BIBentrySTDinterwordspacing

\bibitem{pfaff2025scalablereal2simphysicsawareasset}
\BIBentryALTinterwordspacing
N.~Pfaff, E.~Fu, J.~Binagia, P.~Isola, and R.~Tedrake, ``Scalable real2sim: Physics-aware asset generation via robotic pick-and-place setups,'' 2025. [Online]. Available: \url{https://arxiv.org/abs/2503.00370}
\BIBentrySTDinterwordspacing

\bibitem{marcucci2022motionplanningobstaclesconvex}
\BIBentryALTinterwordspacing
T.~Marcucci, M.~Petersen, D.~von Wrangel, and R.~Tedrake, ``Motion planning around obstacles with convex optimization,'' 2022. [Online]. Available: \url{https://arxiv.org/abs/2205.04422}
\BIBentrySTDinterwordspacing

\bibitem{chen2021generalinhandobjectreorientation}
\BIBentryALTinterwordspacing
T.~Chen, J.~Xu, and P.~Agrawal, ``A system for general in-hand object re-orientation,'' 2021. [Online]. Available: \url{https://arxiv.org/abs/2111.03043}
\BIBentrySTDinterwordspacing

\bibitem{tobin2017domainrandomizationtransferringdeep}
\BIBentryALTinterwordspacing
J.~Tobin, R.~Fong, A.~Ray, J.~Schneider, W.~Zaremba, and P.~Abbeel, ``Domain randomization for transferring deep neural networks from simulation to the real world,'' 2017. [Online]. Available: \url{https://arxiv.org/abs/1703.06907}
\BIBentrySTDinterwordspacing

\bibitem{fey2025bridgingsimtorealgapathletic}
\BIBentryALTinterwordspacing
N.~Fey, G.~B. Margolis, M.~Peticco, and P.~Agrawal, ``Bridging the sim-to-real gap for athletic loco-manipulation,'' 2025. [Online]. Available: \url{https://arxiv.org/abs/2502.10894}
\BIBentrySTDinterwordspacing

\bibitem{nasiriany2024robocasalargescalesimulationeveryday}
\BIBentryALTinterwordspacing
S.~Nasiriany, A.~Maddukuri, L.~Zhang, A.~Parikh, A.~Lo, A.~Joshi, A.~Mandlekar, and Y.~Zhu, ``Robocasa: Large-scale simulation of everyday tasks for generalist robots,'' 2024. [Online]. Available: \url{https://arxiv.org/abs/2406.02523}
\BIBentrySTDinterwordspacing

\bibitem{li2021universalrepresentationlearningmultiple}
\BIBentryALTinterwordspacing
W.-H. Li, X.~Liu, and H.~Bilen, ``Universal representation learning from multiple domains for few-shot classification,'' 2021. [Online]. Available: \url{https://arxiv.org/abs/2103.13841}
\BIBentrySTDinterwordspacing

\bibitem{theorydifferentdomains}
\BIBentryALTinterwordspacing
S.~Ben-David, J.~Blitzer, K.~Crammer, A.~Kulesza, F.~Pereira, and J.~Vaughan, ``A theory of learning from different domains,'' \emph{Machine Learning}, vol.~79, pp. 151--175, 2010. [Online]. Available: \url{http://www.springerlink.com/content/q6qk230685577n52/}
\BIBentrySTDinterwordspacing

\bibitem{xie2023doremioptimizingdatamixtures}
\BIBentryALTinterwordspacing
S.~M. Xie, H.~Pham, X.~Dong, N.~Du, H.~Liu, Y.~Lu, P.~Liang, Q.~V. Le, T.~Ma, and A.~W. Yu, ``Doremi: Optimizing data mixtures speeds up language model pretraining,'' 2023. [Online]. Available: \url{https://arxiv.org/abs/2305.10429}
\BIBentrySTDinterwordspacing

\bibitem{kim2024openvlaopensourcevisionlanguageactionmodel}
\BIBentryALTinterwordspacing
M.~J. Kim, K.~Pertsch, S.~Karamcheti, T.~Xiao, A.~Balakrishna, S.~Nair, R.~Rafailov, E.~Foster, G.~Lam, P.~Sanketi, Q.~Vuong, T.~Kollar, B.~Burchfiel, R.~Tedrake, D.~Sadigh, S.~Levine, P.~Liang, and C.~Finn, ``{OpenVLA}: An open-source vision-language-action model,'' 2024. [Online]. Available: \url{https://arxiv.org/abs/2406.09246}
\BIBentrySTDinterwordspacing

\bibitem{hejna2024remixoptimizingdatamixtures}
\BIBentryALTinterwordspacing
J.~Hejna, C.~Bhateja, Y.~Jian, K.~Pertsch, and D.~Sadigh, ``Re-mix: Optimizing data mixtures for large scale imitation learning,'' 2024. [Online]. Available: \url{https://arxiv.org/abs/2408.14037}
\BIBentrySTDinterwordspacing

\bibitem{song2022denoisingdiffusionimplicitmodels}
\BIBentryALTinterwordspacing
J.~Song, C.~Meng, and S.~Ermon, ``Denoising diffusion implicit models,'' 2022. [Online]. Available: \url{https://arxiv.org/abs/2010.02502}
\BIBentrySTDinterwordspacing

\bibitem{ho2020denoisingdiffusionprobabilisticmodels}
\BIBentryALTinterwordspacing
J.~Ho, A.~Jain, and P.~Abbeel, ``Denoising diffusion probabilistic models,'' 2020. [Online]. Available: \url{https://arxiv.org/abs/2006.11239}
\BIBentrySTDinterwordspacing

\bibitem{he2015deepresiduallearningimage}
\BIBentryALTinterwordspacing
K.~He, X.~Zhang, S.~Ren, and J.~Sun, ``Deep residual learning for image recognition,'' 2015. [Online]. Available: \url{https://arxiv.org/abs/1512.03385}
\BIBentrySTDinterwordspacing

\bibitem{lee2024behaviorgenerationlatentactions}
\BIBentryALTinterwordspacing
S.~Lee, Y.~Wang, H.~Etukuru, H.~J. Kim, N.~M.~M. Shafiullah, and L.~Pinto, ``Behavior generation with latent actions,'' 2024. [Online]. Available: \url{https://arxiv.org/abs/2403.03181}
\BIBentrySTDinterwordspacing

\bibitem{lynch2022interactivelanguagetalkingrobots}
\BIBentryALTinterwordspacing
C.~Lynch, A.~Wahid, J.~Tompson, T.~Ding, J.~Betker, R.~Baruch, T.~Armstrong, and P.~Florence, ``Interactive language: Talking to robots in real time,'' 2022. [Online]. Available: \url{https://arxiv.org/abs/2210.06407}
\BIBentrySTDinterwordspacing

\bibitem{graesdal2024tightconvexrelaxationscontactrich}
\BIBentryALTinterwordspacing
B.~P. Graesdal, S.~Y.~C. Chia, T.~Marcucci, S.~Morozov, A.~Amice, P.~A. Parrilo, and R.~Tedrake, ``Towards tight convex relaxations for contact-rich manipulation,'' 2024. [Online]. Available: \url{https://arxiv.org/abs/2402.10312}
\BIBentrySTDinterwordspacing

\bibitem{aydinoglu2024consensuscomplementaritycontrolmulticontact}
\BIBentryALTinterwordspacing
A.~Aydinoglu, A.~Wei, W.-C. Huang, and M.~Posa, ``Consensus complementarity control for multi-contact mpc,'' 2024. [Online]. Available: \url{https://arxiv.org/abs/2304.11259}
\BIBentrySTDinterwordspacing

\bibitem{kang2025globalcontactrichplanningsparsityrich}
\BIBentryALTinterwordspacing
S.~Kang, G.~Liu, and H.~Yang, ``Global contact-rich planning with sparsity-rich semidefinite relaxations,'' 2025. [Online]. Available: \url{https://arxiv.org/abs/2502.02829}
\BIBentrySTDinterwordspacing

\bibitem{tedrake2019drake}
\BIBentryALTinterwordspacing
R.~Tedrake and the Drake Development~Team, ``Drake: Model-based design and verification for robotics,'' 2019. [Online]. Available: \url{https://drake.mit.edu}
\BIBentrySTDinterwordspacing

\bibitem{zhao2024alohaunleashedsimplerecipe}
\BIBentryALTinterwordspacing
T.~Z. Zhao, J.~Tompson, D.~Driess, P.~Florence, K.~Ghasemipour, C.~Finn, and A.~Wahid, ``Aloha unleashed: A simple recipe for robot dexterity,'' 2024. [Online]. Available: \url{https://arxiv.org/abs/2410.13126}
\BIBentrySTDinterwordspacing

\bibitem{zhao2022representationlearningvideos}
T.~Z. Zhao, S.~Karamcheti, T.~Kollar, C.~Finn, and P.~Liang, ``What makes representation learning from videos hard for control?'' RSS Workshop on Scaling Robot Learning, 2022.

\bibitem{qureshi2024splatsimzeroshotsim2realtransfer}
\BIBentryALTinterwordspacing
M.~N. Qureshi, S.~Garg, F.~Yandun, D.~Held, G.~Kantor, and A.~Silwal, ``Splatsim: Zero-shot sim2real transfer of rgb manipulation policies using gaussian splatting,'' 2024. [Online]. Available: \url{https://arxiv.org/abs/2409.10161}
\BIBentrySTDinterwordspacing

\bibitem{rao2020rlcycleganreinforcementlearningaware}
\BIBentryALTinterwordspacing
K.~Rao, C.~Harris, A.~Irpan, S.~Levine, J.~Ibarz, and M.~Khansari, ``Rl-cyclegan: Reinforcement learning aware simulation-to-real,'' 2020. [Online]. Available: \url{https://arxiv.org/abs/2006.09001}
\BIBentrySTDinterwordspacing

\bibitem{kaplan2020scalinglawsneurallanguage}
\BIBentryALTinterwordspacing
J.~Kaplan, S.~McCandlish, T.~Henighan, T.~B. Brown, B.~Chess, R.~Child, S.~Gray, A.~Radford, J.~Wu, and D.~Amodei, ``Scaling laws for neural language models,'' 2020. [Online]. Available: \url{https://arxiv.org/abs/2001.08361}
\BIBentrySTDinterwordspacing

\bibitem{edwards2016censoringrepresentationsadversary}
\BIBentryALTinterwordspacing
H.~Edwards and A.~Storkey, ``Censoring representations with an adversary,'' 2016. [Online]. Available: \url{https://arxiv.org/abs/1511.05897}
\BIBentrySTDinterwordspacing

\bibitem{goodfellow2014generativeadversarialnetworks}
\BIBentryALTinterwordspacing
I.~J. Goodfellow, J.~Pouget-Abadie, M.~Mirza, B.~Xu, D.~Warde-Farley, S.~Ozair, A.~Courville, and Y.~Bengio, ``Generative adversarial networks,'' 2014. [Online]. Available: \url{https://arxiv.org/abs/1406.2661}
\BIBentrySTDinterwordspacing

\bibitem{JMLR:v13:gretton12a}
\BIBentryALTinterwordspacing
A.~Gretton, K.~M. Borgwardt, M.~J. Rasch, B.~Sch{{\"o}}lkopf, and A.~Smola, ``A kernel two-sample test,'' \emph{Journal of Machine Learning Research}, vol.~13, no.~25, pp. 723--773, 2012. [Online]. Available: \url{http://jmlr.org/papers/v13/gretton12a.html}
\BIBentrySTDinterwordspacing

\bibitem{zhao2023learningfinegrainedbimanualmanipulation}
\BIBentryALTinterwordspacing
T.~Z. Zhao, V.~Kumar, S.~Levine, and C.~Finn, ``Learning fine-grained bimanual manipulation with low-cost hardware,'' 2023. [Online]. Available: \url{https://arxiv.org/abs/2304.13705}
\BIBentrySTDinterwordspacing

\end{thebibliography}

\appendix
\subsection{Real-World Evaluation}
\label{appendix:real}
\begin{figure}[h]
    \centering
    \begin{subfigure}[b]{0.59\linewidth}
        \centering
        \includegraphics[width=0.925\linewidth]{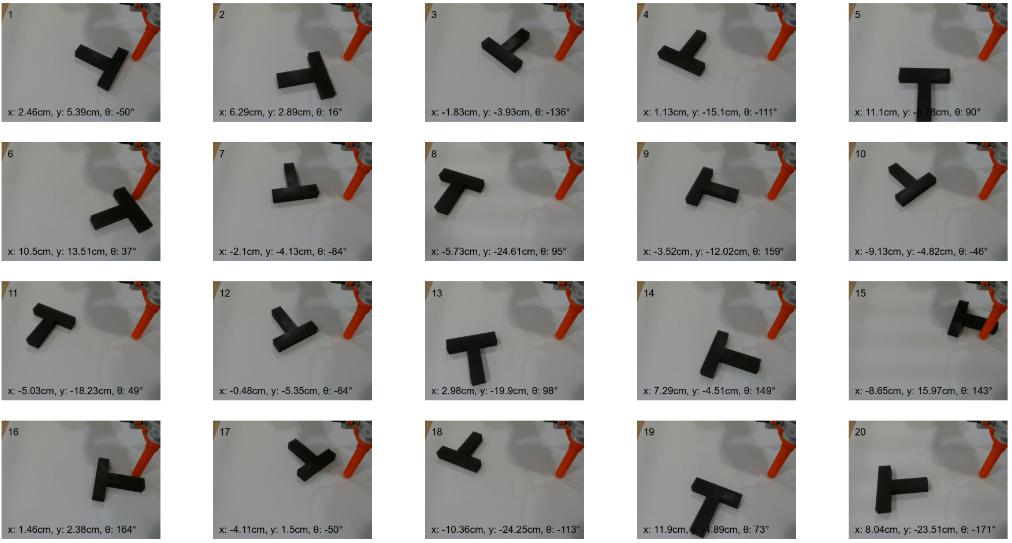}
        \caption{}
    \end{subfigure}
    \hfill
    \begin{subfigure}[b]{0.39\linewidth}
        \centering
        \includegraphics[width=\linewidth]{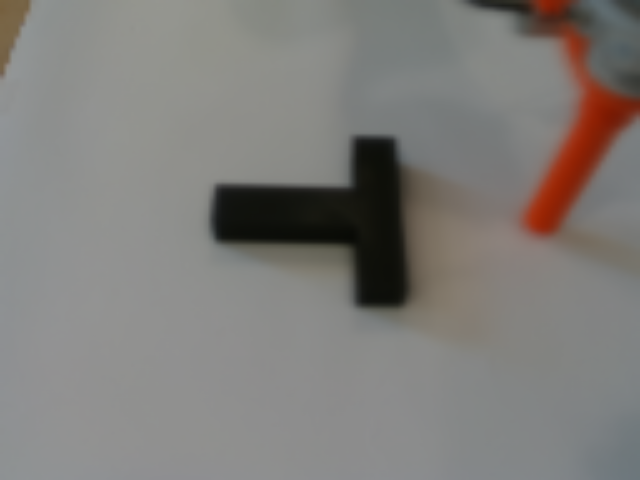}
        \caption{}
    \end{subfigure}
    \caption{a) Initial conditions for the real-world trials. b) An overlay of the final state for the 160 successful demos. We use the same success criteria to evaluate the policy rollouts. The overlay is fairly noiseless, illustrating the strictness of our evaluations.\vspace{-1\baselineskip}}
    \label{fig:appendix}
\end{figure}

For each policy, we report the performance of the best checkpoint over the 20 initial conditions in Figure \ref{fig:appendix}. A trial is successful if a human teleoperator would not readjust the final slider pose. The time limit is 90s. We plot standard error (SE) bars, calculated as $\sqrt{p(1 - p)/n}$, where $p$ is the empirical success rate and $n$ is the number of trials. 

\subsection{Simulation Evaluation}
\label{appendix:sim}
A trial is successful if the robot returns to its default position and the slider's pose is within 1.5cm and $3.5^\mathrm o$ of the goal. The time limit is 75s. For each policy, we evaluate all checkpoints for up to 200 trials. During evaluation, we periodically discard low performing checkpoints according to a Bayesian framework to save compute. We report the performance of the best checkpoints and compute SE bars. Both our \textit{target sim} evaluation set up and simulated data generation scripts are available here: \url{https://github.com/sim-and-real-cotraining/planning-through-contact}.

\subsection{Training Details}
\label{appendix:training}
Policies were trained using a fork of the code provided by \cite{chi2024diffusionpolicy}. We re-used most hyperparameters but performed a small sweep to select the batch size, learning rate, and number of optimizer steps. We used their U-Net architecture with a ResNet18 backbone \cite{he2015deepresiduallearningimage} and trained end-to-end.

To ensure a fair comparison, we trained each model for approximately the same number of optimizer steps and the same hyperparameters. We saved 4-5 checkpoints along the way. For our finetuning experiments, we reduced the learning rate by 10x and reported performance for the best checkpoint. The configuration files for all training runs are available here: \url{https://github.com/sim-and-real-cotraining/diffusion-policy}.

\end{document}